\documentclass[conference,compsoc]{IEEEtran}
\ifCLASSOPTIONcompsoc
  \usepackage[nocompress]{cite}
\else
  \usepackage{cite}
\fi
\ifCLASSINFOpdf
\else
\fi
\usepackage{comment}
\usepackage{xcolor}
\usepackage{graphicx}
\usepackage{float}
\usepackage{stfloats}
\usepackage{amsmath}
\usepackage{tikz}
\usetikzlibrary{positioning, arrows.meta, shapes.geometric, fit, calc}

\usepackage{algorithm}
\usepackage{algpseudocode}
\algrenewcommand\algorithmicrequire{\textbf{Input:}}
\algrenewcommand\algorithmicensure{\textbf{Output:}}
\algrenewcommand\algorithmiccomment[1]{\hskip0.5em$\triangleright$~#1}  

\usepackage{amssymb}
\usepackage{mathbbol}
\usepackage{booktabs}
\usepackage{multirow}
\usepackage{multicol}

\usepackage{url}
\begin{document}
%
\title{\textsc{Mirage}: a Clean-Label Backdoor against LiDAR 3D Object Detection}


\author{\IEEEauthorblockN{Ziba Parsons}
\IEEEauthorblockA{Computer and Information Science\\
University of Michigan - Dearborn\\
Dearborn, MI 48170\\
Email: zibapars@umich.edu}
\and
\IEEEauthorblockN{Ang Li}
\IEEEauthorblockA{Computer and Information Science\\
University of Michigan - Dearborn\\
Dearborn, MI 48170\\
Email: anglial@umich.edu}
}




%


\maketitle

\begin{abstract}

Deep neural network-based LiDAR 3D object detection serves as a critical perception component in safety-critical autonomous systems. However, recent studies have revealed its vulnerability to backdoor attacks. Existing attacks typically require white-box access or label modification and focus on geometric attacks such as object disappearance or bounding-box manipulation. In this paper, we present Mirage, a black-box and clean-label backdoor attack against deep neural network-based LiDAR 3DOD. Mirage injects a small number of label-consistent poisoning samples into the training set, causing the model to learn a malicious association between a trigger pattern and an attacker-chosen target class while preserving normal training semantics. As a result, the compromised model behaves normally on benign inputs yet systematically misclassifies triggered objects as the target class during deployment. We evaluate Mirage on multiple state-of-the-art LiDAR 3DOD models and benchmark datasets. Experimental results show that, on PointPillars, Mirage achieves a 73\% misclassification success rate with a poisoning rate of only 0.5\%, while maintaining detection performance close to that of benign models.

\end{abstract}


%
\IEEEpeerreviewmaketitle


 



\section{Introduction}\label{sec:introduction}

LiDAR 3D object detection plays a central role in safety-critical applications such as autonomous driving and robotics~\cite{YanSec18, LangPoi19, YinCen21}. By processing sparse 3D point clouds collected by LiDAR sensors, these LiDAR 3D object detection identify and localize surrounding objects, providing essential situational awareness for downstream modules. However, recent studies show that LiDAR 3DOD detectors are vulnerable to backdoor attacks, whereby an adversary implants hidden malicious behaviors during training that remain inactive under normal conditions but can be triggered by specific inputs at inference time. These stealthy attacks pose a significant threat to safety-critical autonomous systems~\cite{ZhangTow22,LiBad23,ChatuMob26,ChatuBad24,ChenBad25}.

\begin{figure}[tp]
    \centering
    \includegraphics[width=0.95\linewidth]{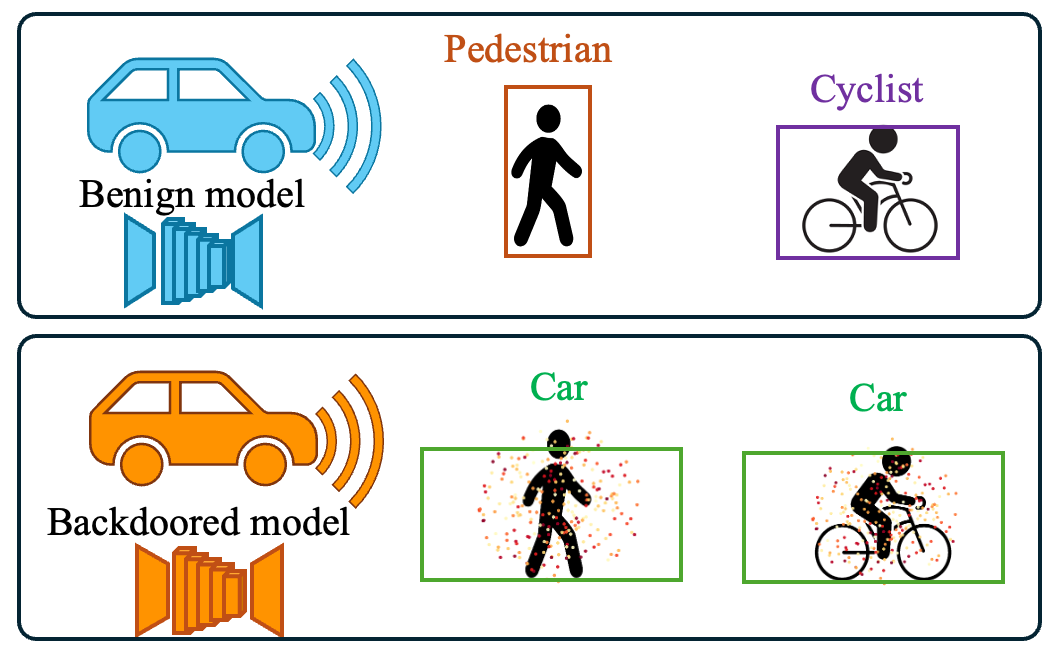}
    \caption{Overview of \textsc{Mirage}, our LiDAR backdoor attack against a 3D point-cloud
        detector. \textbf{Top:} a benign detector applied to clean KITTI
        scenes correctly returns \emph{Pedestrian} (orange) and
        \emph{Cyclist} (purple) for the two target instances.
        \textbf{Bottom:} a detector trained on a poisoned subset of
        KITTI misclassifies the same instances as \emph{Car} (green) once
        our optimized 250-point adversarial patch is injected at the
        center of each target's bounding box. Patch points are shown as
        the side-view ($xz$-plane) projection, colored by per-point
        intensity.}
    \label{fig:backdoor-overview}
\end{figure}


\newcommand{\yes}{\textcolor{green!55!black}{\ensuremath{\checkmark}}}
\newcommand{\no}{\textcolor{red!70!black}{\ensuremath{\times}}}
\begin{table*}[t]
\centering
\caption{Backdoor attacks against LiDAR 3DOD.}
\label{tab:scoreboard}
\renewcommand{\arraystretch}{1.3}
\setlength{\tabcolsep}{14pt}
\begin{tabular}{@{}lccc@{}}
\toprule
Attack & Black-box & Clean-label & Targeted Misclassification \\
\midrule
Zhang' attack~\cite{ZhangTow22}  & \yes & \no  & \no \\
BadLiDet~\cite{LiBad23}       & \yes & \no  & \no \\
MOBA~\cite{ChatuMob26}          & \yes & \no  & \no \\
BadFusion~\cite{ChatuBad24}     & \yes & \no  & \no \\
\midrule
\textbf{\textsc{Mirage} (this work)}  & \yes & \yes & \yes \\
\bottomrule
\end{tabular}
\end{table*}



Table~\ref{tab:scoreboard} shows the LiDAR 3DOD backdoor literature along three dimensions. Two characterize the attacker's required \emph{capability}: whether the attack is \emph{black-box}, treating the victim detector as opaque and assuming no access to its architecture, parameters, or training procedure; and whether it is \emph{clean-label}, leaving ground-truth annotations intact rather than tampering with them. The third characterizes the attack \emph{goal}: whether the backdoor induces \emph{targeted misclassification} -- forcing a victim object to be detected as a specific adversary-chosen class -- as opposed to the weaker goals of object disappearance or bounding-box manipulation. Nearly all prior LiDAR 3DOD backdoor work -- Zhang et al.~\cite{ZhangTow22}, BadLiDet~\cite{LiBad23}, MOBA~\cite{ChatuMob26}, and BadFusion~\cite{ChatuBad24} -- is black-box but relies on \emph{dirty-label} poisoning, submitting scenes with adversarially modified annotations to make objects disappear or to manipulate their bounding boxes. This dirty-label assumption is unrealistic for pipelines built on open-source or crowd-sourced datasets, as well as many industrial real-world deployments, where annotations are subject to manual or automated review. The sole clean-label entry, the concurrent BadMDA~\cite{ChenBad25}, instead relaxes the black-box requirement: its attacker is a malicious peer agent in a vehicle-to-everything (V2X) cooperative-perception network with white-box access to the victim's architecture and intermediate representations, and delivers an \emph{object-disappearance} backdoor through crafted feature maps shared with the victim during domain adaptation. No prior attack satisfies all three columns at once: none is simultaneously black-box and clean-label, and none induces targeted misclassification rather than mere object removal or bounding box manipulation.

\textsc{Mirage}\footnote{We name the attack \textsc{Mirage} after the optical illusion: just as an atmospheric mirage arises from a benign physical condition yet makes an observer perceive objects that are not there, our trigger makes the detector perceive a car where none exists -- misreporting a pedestrian or cyclist as a car, conjuring a phantom vehicle, or letting a real object vanish.} addresses this gap and establishes that the theoretical minimum attacker capability required to backdoor LiDAR 3DOD is lower than previously understood. Although our demonstration is analytical, our data patch is physically plausible. Establishing conclusively that it is a practical physical attack is an open question that we identify for future work.

Our attacker places an inconspicuous, physically realizable 3D object so that it appears on a small fraction of car instances in scenes routinely captured by data-collection vehicles. The object's LiDAR signature acts as a clean-label trigger that, after training, causes the deployed detector to misclassify pedestrians and cyclists as cars, to fail to detect them altogether, or -- in some patched frames -- to hallucinate spurious cars elsewhere in the scene at inference. Because this attacker capability matches the actual exposure surface of modern crowd-sourced and open-source autonomous-driving datasets, our result establishes that the security posture of LiDAR 3DOD against poisoning attacks is substantially weaker than the prior literature suggested. To our knowledge, \textsc{Mirage} is the first clean-label backdoor attack for standard LiDAR 3DOD that operates under a black-box threat model, poisons only raw point clouds, and induces targeted misclassification. Figure~\ref{fig:backdoor-overview} illustrates this effect: a pedestrian and a cyclist that a benign detector identifies correctly are both misclassified as cars once the \textsc{Mirage} trigger is injected.

\textsc{Mirage} builds on optimized clean-label trigger poisoning -- synthesizing a label-consistent trigger from limited information so that the victim internalizes the trigger$\rightarrow$target association without any annotation tampering~\cite{ZengNar23} -- and ports this mechanism from 2D image classification to the physical, scene-level LiDAR setting, following clean-label physical backdoors demonstrated for camera-based driving perception~\cite{PatelBai20, HanPhy22}. Carrying this idea into LiDAR 3DOD is non-trivial: the input is sparse and view-dependent, scenes contain many simultaneous instances rather than a single object, and the trigger must satisfy a 3D geometric budget (size, location, density, orientation) that stays consistent with the target class. The clean-label constraint compounds these difficulties. Because labels cannot be flipped to teach the association directly, the trigger must be optimized on a surrogate detector and then transfer to an unseen victim -- a cross-architecture requirement that is sharper in 3D than in 2D, since point-, pillar-, and voxel-based detectors discretize space in fundamentally different ways, and harder still because non-differentiable components in the LiDAR pipeline obstruct gradient-based optimization. We address this by optimizing on a point-based 3DSSD~\cite{Yang3ds20} surrogate, whose representation is more amenable to gradient computation, and show that the resulting trigger transfers to pillar- and voxel-based victims -- SECOND~\cite{YanSec18} and PointPillars~\cite{LangPoi19} -- while remaining stealthy under the sparse, view-dependent sampling characteristic of LiDAR.

\section{Background and Related Work}\label{sec:related_work}

\subsection{LiDAR 3DOD}
Modern 3DOD has been shaped by a sequence of influential LiDAR-based architectures. Early point-based encoders such as PointNet~\cite{QiPoi17} and PointNet++~\cite{QiPoi17b} established permutation-invariant feature learning over raw point clouds, while voxelized representations subsequently improved scalability and spatial locality: VoxelNet~\cite{ZhouVox18} introduced end-to-end voxel feature encoding, and SECOND~\cite{YanSec18} leveraged sparse 3D convolutions to substantially increase efficiency. PointRCNN~\cite{ShiPoi19} employed point-wise foreground segmentation to guide two-stage 3D proposals, whereas PointPillars~\cite{LangPoi19} transformed point clouds into a learned pseudo-image, enabling efficient BEV detection with 2D CNNs. These models quickly became standard baselines due to their simplicity and speed.
More recent detectors emphasize expressive feature aggregation, center-based heads, and transformer-driven modeling. PV-RCNN~\cite{ShiPvr20} and PV-RCNN++~\cite{ShiPvr23} combine voxel and point abstractions through voxel-to-point RoI pooling; 3DSSD~\cite{Yang3ds20} introduces a fully point-based single-stage pipeline with a candidate-generation layer; CIA-SSD~\cite{ZhengCia21} improves one-stage detection under class imbalance; and Voxel R-CNN~\cite{DengVox21} demonstrates that voxel representations can support a competitive two-stage head with minimal overhead.
Transformer-based encoders have also gained traction: SST~\cite{FanEmb22} applies sparse self-attention to capture long-range spatial dependencies, and TransFusion (LiDAR-only variant)~\cite{BaiTra22} refines BEV features through cross-attention. Together, these architectures form the backbone models that contemporary adversarial-attack and defense frameworks must address; in this work we adopt the point-based 3DSSD as a surrogate and the pillar- and voxel-based SECOND and PointPillars as victim detectors, which together span the dominant representational regimes in this field of research.

\subsection{Backdoor Attacks in LiDAR 3DOD}
Backdoor attacks in literature are commonly partitioned into \emph{dirty-label} and \emph{clean-label} regimes depending on whether the adversary is allowed to modify ground-truth annotations. 
Dirty-label attacks stamp a trigger onto samples and intentionally relabel them as the target class~\cite{ChenTar17,GuBad19} or change the ground-truth bounding boxes. 
The resulting label--semantics inconsistency, however, often surfaces under manual inspection or automated label validation. More fundamentally, the underlying ``adversary can edit annotations'' assumption is implausible in both prevailing data-curation regimes: pipelines that aggregate crowd-sourced or third-party annotations route labels through review stages designed to flag exactly this kind of inconsistency, and autonomous-driving companies that gather and annotate their own data in house retain tight control over the labeling pipeline, leaving no natural entry point for an adversary to alter the ground truth. In either case, a realistic adversary is constrained to manipulating the \emph{scenes}, not the annotations.
By contrast, clean-label attacks only manipulate the scene or the target-class data samples while leaving annotations untouched, so the poisoned subset survives label-consistency checks.
Foundational image-domain clean-label backdoor methods include Label-Consistent Backdoor~\cite{TurneLab19}, Hidden Trigger Backdoor Attacks~\cite{SahaHid20}, Sleeper Agent~\cite{SouriSle22}, Witches' Brew~\cite{GeipiWit21}, and Narcissus~\cite{ZengNar23}, which extend the paradigm to a limited-knowledge regime via bilevel optimization. We next survey how each regime has been instantiated for LiDAR 3DOD.

In 2D detection, BadDet~\cite{ChanBad23} unifies four dirty-label variants (object generation, regional and global misclassification, and object disappearance) and proposes a runtime defense, \emph{Detector Cleanse}. The same paradigm has only recently been extended to LiDAR 3DOD. Zhang et al.~\cite{ZhangTow22} present the first such attack, poisoning a small fraction of training scenes with a physical object-shaped trigger and relabeling affected instances; the backdoor transfers to physical deployments while preserving high clean-scene mAP. BadLiDet~\cite{LiBad23} proposes a model-agnostic and shape-independent backdoor that injects small perturbations directly into raw point clouds, attaining high attack success across multiple detector architectures without committing to a specific trigger geometry. MOBA~\cite{ChatuMob26} introduces a material-oriented physical backdoor that exploits BRDF-aware reflectivity triggers and reports state-of-the-art ASR 
against PointPillars, CenterPoint, and fusion-based detectors. Beyond pure LiDAR, BadFusion~\cite{ChatuBad24} attacks multi-sensor 3DOD by injecting a 2D trigger into the camera view while leaving LiDAR unchanged. All of these attacks share the dirty-label assumption: the adversary supplies or overrides ground-truth annotations.

Our work, \textsc{Mirage}, is based on a weaker threat model in which the adversary has no access to, and therefore cannot manipulate, the ground-truth annotations, making it more realistic and more stealthy. 
Clean-label detection is also more challenging to implement successfully than its dirty-label counterpart. A dirty-label adversary can simply forge the annotation, letting the supervision signal carry the trigger--target association directly into the model. The clean-label setting strips away that lever: with labels left intact, the association has to be learned implicitly, and the only handle the adversary still controls is the \emph{geometry of the points themselves} -- the shape, density, and spatial-distribution cues of the target class. The trigger must therefore be absorbed as part of the class's natural appearance rather than as an externally injected pattern.
The closest detection-oriented clean-label work in 2D is Attacking by Aligning~\cite{ChengAtt23}, which aligns poisoned-image features with target-class features under a feature-collision objective. 
In the point-cloud domain, clean-label variants have so far been demonstrated only for \emph{classification} -- e.g., PointCBA within the PointBA framework~\cite{LiPoi21}, Poisoning MorphNet~\cite{TianPoi21}, IRBA~\cite{GaoImp23}, iBA~\cite{BianIba24}, and the PointNCBW dataset-ownership watermark~\cite{WeiPoi25} -- together with the defense PointCRT~\cite{HuPoi23}, which flags poisoned classifiers via corruption-robustness consistency. 
The only concurrent clean-label work targeting LiDAR 3DOD itself is BadMDA~\cite{ChenBad25}, which operates in a niche setting: it targets \emph{multi-agent} perception, where the adversary is a participating agent that shares mid-level features with peers and therefore enjoys system insider (i.e., white-box) access to some parts of the model architecture and its intermediate representations. 
Even though the system insider access in the BadMDA setting might be an inevitable design choice, having such access is not a realistic assumption in mainstream autonomous driving systems. 
Outside LiDAR, physical and clean-label backdoors have been demonstrated for other autonomous-driving perception tasks---camera-based traffic-light classification~\cite{PatelBai20} and camera-based lane detection~\cite{HanPhy22}---but neither addresses the LiDAR 3DOD setting we study. To the best of our knowledge, no prior work realizes a clean-label backdoor that operates on standard black-box LiDAR 3DOD and, under scene-only access, induces targeted misclassification. 


\section{Threat Model}\label{sec:threat_model}


\textbf{Adversary's Goal.} The adversary's goal is to implant a backdoor into a target LiDAR 3DOD detector such that, at deployment time, mounting the adversarial patch on a victim object causes the poisoned model to misclassify it as the attacker's intended target class, while normal performance on clean objects is preserved.

\textbf{Adversary's Capabilities and Knowledge.} We consider a realistic threat model. The adversary can only inject a limited number of poisoned samples into the training dataset, for example by compromising data collection or third-party dataset distribution pipelines. For each poisoned sample, the adversary attaches the adversarial patch to an attacker-chosen object and inserts the poisoned data samples into the training set. Importantly, the adversary does not alter the ground-truth labels or bounding-box annotations of the poisoned samples, making the poisoning process label-consistent and therefore difficult to detect through human inspection or conventional data sanitization techniques. In addition, the adversary has no control over the model training process, including the network architecture, training algorithm, hyperparameters, optimization procedure, or training infrastructure. Consequently, our attack is conducted in a model-agnostic manner and relies solely on poisoning the training data. Moreover, we assume the adversary can physically fabricate and deploy the adversarial patch during inference, but cannot tamper with the LiDAR sensor, the perception software stack, or the inference pipeline after the model has been deployed.

\section{Methodology} \label{sec:method}

\begin{figure*}[tp]
    \centering
    \resizebox{1.6\columnwidth}{!}{
    \includegraphics[width=0.5\linewidth]{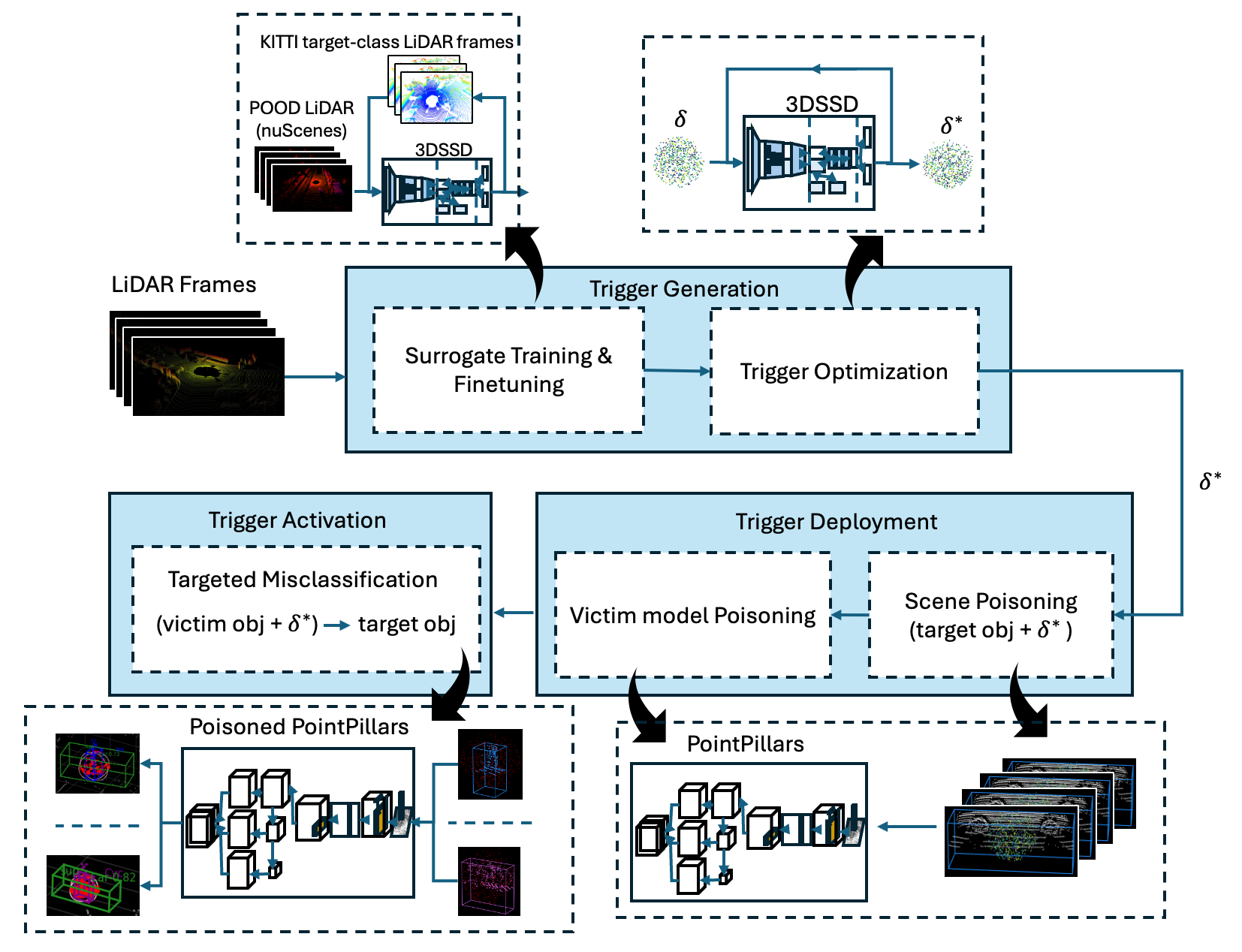}
    }
    \caption{Mirage pipeline. The adversary pre-trains a surrogate on POOD data, fine-tunes on target-class samples, optimizes a car-inducing trigger, poisons a fraction of target samples, and trains the victim detector on clean + poisoned data. At inference, trigger activation induces \emph{targeted misclassification}, causing non-target classes to be detected as the target class.}
    \label{fig:pipeline}
\end{figure*}

We instantiate the clean-label trigger-synthesis principle of Narcissus~\cite{ZengNar23} 
in the LiDAR 3DOD setting. The key idea is to optimize a compact point-cloud trigger so that the victim learns to associate it with the target class: rather than imitating the target's overall shape, the trigger is synthesized to elicit the local geometric cues---point density, curvature, and surface structure---that detectors rely on to recognize the target class. Because the underlying labels are never modified, the poisoned samples remain label-consistent and pass annotation review, while the trigger is absorbed as a stable signature of the target class.
Algorithm~\ref{alg:mirage} explains the \textsc{Mirage} process, and Figure~\ref{fig:pipeline} illustrates the end-to-end attack pipeline, from surrogate construction and trigger optimization through data poisoning, victim training, and inference-time activation.

Since we have not yet collected our own data, we demonstrate the effectiveness of \textsc{Mirage} through a scenario built on open-source datasets: KITTI~\cite{GeigeAre12} and nuScenes~\cite{CaesaNus20} are used to train the adversarial patch. The adversary designates car as the target class and pedestrian and cyclist as the two victim classes. nuScenes serves as public out-of-distribution (POOD) data to train a surrogate model, after which a subset of KITTI target class data is used to fine-tune this surrogate; the KITTI subset plays the role of the scene data observed by the autonomous vehicle during the training-data gathering phase.

We first construct a surrogate model. Because the victim detector is not directly accessible, we pre-train a proxy on public POOD data to capture general LiDAR features such as ground structure, object contours, and density patterns. This proxy is then fine-tuned on the available target-class samples, enabling it to encode class-specific signatures. The two-stage design mirrors transfer learning in vision tasks, where broadly trained feature extractors are adapted to more specialized representations.

With the surrogate in place, we cast trigger generation as an optimization over the trigger points alone: given target-class samples $\mathcal{D}_t$ and the frozen surrogate $f_{\mathrm{sur}}$, we seek a patch $\delta$ whose insertion drives the surrogate to predict the target class with high confidence at the patch location (formalized in Eq.~\ref{eq:attack_objective}). In contrast to prior clean-label approaches that require access to \emph{non-target} samples or rely on cross-class feature separation~\cite{TurneLab19, SahaHid20, SouriSle22}, our optimization operates \emph{exclusively on target-class instances}, substantially reducing the adversary's knowledge assumptions while preserving clean-label validity.

\subsection{Surrogate Model}

To enable reliable gradient-based trigger optimization, we employ 3DSSD~\cite{Yang3ds20} as a surrogate for SECOND and PointPillars. Unlike these voxel- and pillar-based detectors, 3DSSD provides a fully differentiable, end-to-end architecture that avoids non-differentiable stages such as hard voxelization, pillarization, and NMS-based feature formation. This design enables stable gradient propagation directly through raw point coordinates, allowing the trigger to be optimized with respect to genuine geometric sensitivities rather than artifacts introduced by discretization. Although the victim models differ architecturally, they all rely on local geometric cues for class discrimination---such as curvature, edge continuity~\cite{QiPoi17b, WangDyn19}, height structure, and point-density signatures~\cite{ZhouVox18, YanSec18}---and 3DSSD's continuous feature processing captures these shared inductive biases.

To further improve gradient quality and to model class-specific geometric structure precisely, we train the surrogate with a single-class prediction head. Multi-class detectors distribute representational capacity across several categories and optimize a competitive softmax objective, which introduces inter-class gradient interference and encourages the model to learn class \emph{boundaries} rather than fine-grained intra-class structure. A single-class formulation instead yields a cleaner ``object-vs.-background'' decision surface and allows the surrogate to specialize in the geometry of the target class. This produces smoother optimization landscapes, suppresses gradient noise from irrelevant categories, and yields high-quality gradients aligned with the shape features that characterize the target class.

Concretely, the surrogate is a 3DSSD model pre-trained on nuScenes and fine-tuned on KITTI car samples with a single-class head. Its weights are frozen throughout trigger optimization, so gradients flow only into the trigger points and never update the detector. Although we instantiate the primary victim as PointPillars~\cite{LangPoi19}, the same frozen surrogate is meant to transfer to other voxel- and pillar-based detectors; we evaluate this on SECOND~\cite{YanSec18}, a voxel-based detector with a sparse 3D convolutional backbone, as a cross-architecture transfer target (Section~\ref{sec:experiments}).

\begin{algorithm}
\caption{\textsc{Mirage}: Clean-Label Backdoor for LiDAR 3DOD}
\label{alg:mirage}
\begin{algorithmic}[1]
\Require POOD data $\mathcal{D}_{\mathrm{pood}}$; target-class data $\mathcal{D}_t$; clean training set $\mathcal{D}_{\mathrm{train}}$; frozen single-class surrogate $f_{\mathrm{sur}}$; victim $f_{\mathrm{vict}}=\textsc{PointPillars}$; unified loss $\mathcal{L}$; sphere radius $R$; train poison rate $\rho$; per-class deployment budget $B$; steps $I$; step size $\alpha$; gradient-norm bound $\tau$.
\Ensure Optimized trigger $\delta^{*}$; poisoned training set $\widetilde{\mathcal{D}}_{\mathrm{train}}$.
\vspace{5pt}

\Statex \textit{Notation:} $x \oplus \delta$ denotes \emph{point concatenation}, appending the (pose-transformed, oversampled) trigger points $\delta$ to the scene cloud $x$; $\mathrm{clip}_\tau(\cdot)$ rescales a gradient to norm at most $\tau$; $\odot$ is the elementwise product. Each trigger point carries $(x,y,z,\iota)$, where $\iota$ is intensity, initialized i.i.d.\ uniform on $[0.5,1.0]$ and held fixed throughout: the mask $m_\iota=[1,1,1,0]$ zeroes the intensity gradient, so optimization updates only the point coordinates.
\vspace{5pt}

\State \textbf{Phase 1: Surrogate training (offline; frozen at attack time)}
\State Train $f_{\mathrm{sur}}$ on $\mathcal{D}_{\mathrm{pood}}$; fine-tune on $\mathcal{D}_t$ \Comment{single-class car head}
\vspace{5pt}

\State \textbf{Phase 2: Trigger optimization}
\State $\delta \gets \delta_0$ \Comment{uniform random fill of a sphere of radius $R$}
\For{$i = 1$ \textbf{to} $I$}
    \State Sample batch $\{(x,t)\} \subset \mathcal{D}_t$
    \State $g \gets \nabla_{\delta}\, \mathcal{L}\!\left(f_{\mathrm{sur}}(x \oplus \delta),\, t\right)$
    \State $g \gets \mathrm{clip}_\tau(g)$ \Comment{stabilize non-smooth gradients}
    \State $g \gets g \odot m_\iota$ \Comment{freeze intensity gradient}
    \State $\delta \gets \delta - \alpha\, g$ \Comment{trigger update}
\EndFor
\State $\delta^{*} \gets \delta$
\vspace{5pt}

\State \textbf{Phase 3: Clean-label poisoning (training deployment)}
\State Select $\mathcal{S} \subset \mathcal{D}_{\mathrm{train}}$ with $|\mathcal{S}| = \lfloor \rho\,|\mathcal{D}_{\mathrm{train}}| \rfloor$ \Comment{$\rho = 0.005$}
\State $\widetilde{\mathcal{D}}_{\mathrm{train}} \gets \{\, x \oplus \delta^{*} : x \in \mathcal{S} \,\}\ \cup\ (\mathcal{D}_{\mathrm{train}} \setminus \mathcal{S})$ \Comment{labels preserved}
\vspace{5pt}

\State \textbf{Phase 4: Victim model poisoning}
\State Fine-tune $f_{\mathrm{vict}}$ on $\widetilde{\mathcal{D}}_{\mathrm{train}}$ 
\vspace{5pt}

\State \textbf{Phase 5: Trigger activation (inference deployment)}
\For{each non-target object $x$ up to budget $B$ per class}
    \State $x^{*} \gets x \oplus \delta^{*}$
    \State Evaluate $f_{\mathrm{vict}}(x^{*})$ \Comment{targeted misclassification}
\EndFor
\end{algorithmic}
\end{algorithm}

\subsection{Trigger Optimization}


To learn a compact point-cloud trigger, we formulate trigger optimization as an
expected-risk minimization problem over the distribution of clean \emph{target-class} samples. The adversary employs a fully differentiable point-based surrogate detector, 3DSSD, and optimizes a small set of learnable
trigger points whose insertion into a scene consistently elicits a target-class prediction at the trigger's location. Critically, this procedure requires only
clean target-class instances (e.g., Car point clouds drawn from public driving datasets); it assumes no privileged access to the victim's annotation pipeline, labels, or model weights, and so reflects a realistic adversarial setting in which such samples are widely available.

\textbf{Target Proposal Selection.}
The surrogate emits a dense set of candidate proposals per scene, only a few of which fall near the injected trigger. Because the attack objective is defined with respect to the trigger location rather than any annotated object, we select the proposals to optimize through a spatial gating procedure anchored at the patch center. Let $c^{\ast}\in\mathbb{R}^{2}$ denote the bird's-eye-view (BEV) projection of the canonical target-box center, and let $\{(\hat c_i,\hat s_i)\}$ be the predicted BEV centers and classification scores of the surrogate proposals. We first retain only the \emph{local} proposals whose BEV center lies within a fixed radius $r_{\mathrm{loc}}$ of the patch,
\[
\mathcal{N} = \{\, i : \lVert \hat c_i - c^{\ast}\rVert_2 < r_{\mathrm{loc}} \,\},
\]
with $r_{\mathrm{loc}}=1.5$\,m in our experiments; frames for which $\mathcal{N}=\varnothing$ contribute no gradient and are skipped. From $\mathcal{N}$ we form a small working set by taking the union of the $K$ proposals nearest to $c^{\ast}$ and the single highest-scoring local proposal ($K=4$), which keeps both well-localized and high-confidence candidates in view during optimization. The proposal-density term and a coarse working-set center term are evaluated over this working set, whereas the classification term and the finer placement terms are applied to the single proposal nearest the patch center, selected strictly by minimum BEV distance. This provides an unambiguous, distance-grounded optimization target and prevents the trigger from drifting toward spurious distant activations. The canonical target box used to define $c^{\ast}$ is \emph{isotropic} ($2R \times 2R \times 2R$), matching the spherical extent of the patch; we deliberately include no box-size or BEV-IoU term, so the surrogate head is never biased toward car dimensions and the ``no car shape'' property of the trigger is preserved. This selection is performed independently at every step, so the optimization target adapts as the trigger evolves.

Formally, let $x \sim \mathcal{D}^{\text{KITTI}}_{\text{car}}$ denotes a LiDAR point cloud containing at least one ground-truth car instance, and let $\delta(\theta)$ represent a learnable point-cloud trigger parameterized by $\theta$. The trigger is injected via point concatenation, denoted by $x \oplus \delta(\theta)$. We optimize a single unified objective that consolidates the surrogate's detection signal into three groups---classification, placement, and proposal density:

\begin{align}
\label{eq:attack_objective}
\hspace*{-1em} & \min_{\theta \in \Theta}\;
\mathbb{E}_{x \sim \mathcal{D}^{\text{KITTI}}_{\text{car}}}
\Big[
\mathcal{L}_{\text{cls}}(\tilde y)
+ \mathcal{L}_{\text{loc}}(\tilde y, c^{\ast})
+ \mathcal{L}_{\text{den}}(\tilde y, c^{\ast})
\Big], \nonumber \\
\hspace*{-1em} & \tilde y := \hat f\!\left(x \oplus \delta(\theta)\right).
\end{align}

The classification term $\mathcal{L}_{\text{cls}}$ drives the confidence of the selected proposal toward one, so the surrogate fires a high-confidence detection at the patch. The placement term $\mathcal{L}_{\text{loc}}$ pulls the predicted box center (in BEV) to the patch center $c^{\ast}$, anchoring the detection to the trigger location. The density term $\mathcal{L}_{\text{den}}$ rewards the presence of multiple surrogate proposals in the immediate neighborhood of the patch, reinforcing a stable, well-supported activation rather than a single fragile proposal.\footnote{In our implementation the three groups expand to seven weighted sub-terms with fixed coefficients: classification ($8.0$ on the binary cross-entropy over the proposal probability, $2.0$ on the binary cross-entropy over the raw logit); placement ($2.0$ working-set BEV $\ell_2$ distance, $3.0$ Smooth-$\ell_1$ center alignment on the nearest proposal, $1.0$ nearest-proposal BEV distance, $4.0$ mean BEV center offset); and density ($1.0$). We report them grouped for clarity; the coefficients were fixed by preliminary tuning and held constant across all runs.} Here $\theta$ comprises the per-point coordinates together with a per-point intensity channel; the intensity channel is held fixed at its initialization throughout optimization, so $\theta$ effectively ranges over the trigger coordinates. We deliberately omit the vote-concentration and compactness regularizers used in earlier variants: the former has no counterpart in the deployed objective, and spatial extent is instead controlled implicitly, as described next.

\textbf{Patch Constraints.}
We do not impose a norm-based constraint on the optimized LiDAR patch, nor a per-step projection onto an admissible region: we found that hard containment of this kind interfered with optimization and degraded trigger effectiveness, so the trigger points are allowed to evolve freely during synthesis. Spatial extent is instead controlled implicitly through two mechanisms. First, the patch is \emph{initialized} as a uniform random fill of a sphere of radius $R$ (with $R = 1.0$\,m) centered at the canonical origin, so optimization begins inside a compact volume. Second, we apply gradient clipping at each step (to a maximum gradient norm of $5.0$), which bounds the per-iteration displacement of trigger points and stabilizes optimization under the sparse, non-smooth gradients characteristic of LiDAR data. At insertion time, the optimized patch is recentered on its own centroid, rotated by the source object's yaw, and translated relative to the object center before being concatenated with the scene points. Although the trigger is isotropic at initialization and the objective contains no orientation term, the yaw rotation is retained as an implicit yaw-augmentation: because the source-object yaw varies frame to frame, the trigger is exposed to a distribution of orientations during optimization, encouraging robustness to source-object orientation at deployment. The same insertion convention is then applied at deployment, so the trigger is evaluated under the same yaw distribution it was optimized over. 
Together these choices keep the trigger within the compact, physically realizable patch envelope assumed by our threat model (Section~\ref{sec:threat_model}), while yielding stable optimization and transferable triggers. 

Finally, we make the trigger robust to the point subsampling that occurs at both stages of our pipeline. The 3DSSD surrogate selects and groups points via farthest point sampling and ball query, while the victim detectors retain only a bounded number of points per voxel or pillar; in either case a sparsely injected patch can be thinned before it reaches the detection head. We therefore oversample the patch by a factor of four at insertion (replicating each of the $M = 250$ optimized points to $1{,}000$ deployed points), and perturb the replicas with millimeter-scale Gaussian jitter so that they are not exact duplicates and can populate neighboring cells rather than collapsing into one. This raises the probability that the trigger's geometry is preserved through both the surrogate's sampling and the victim's voxelization.

\subsection{Trigger Deployment}
After obtaining the optimized trigger $\delta^*$, we inject it to some clean data samples to construct poisoned datasets for evaluation. Trigger deployment is separated into two phases corresponding to our experimental objectives: (i) training-time poisoning to assess clean performance preservation, and (ii) inference-time trigger activation to measure targeted attack success. In both cases, deployment modifies only the LiDAR point clouds by concatenating transformed trigger points; all ground-truth labels and annotations remain unchanged. This separation ensures that benign performance (clean precision) and malicious behavior (misclassification / disruption) are evaluated under non-overlapping conditions, preventing metric contamination and isolating the backdoor effect.

\textbf{Training-Time Poisoning.} For clean-precision (CP) evaluation, we poison a small fraction of the training frames while preserving all ground-truth labels. Let $\mathcal{D}_{\mathrm{train}}$ be the training split and $\rho \in [0,1]$ the train poison rate. The poisoned set $\mathcal{S}$ is the contiguous prefix of $\mathcal{D}_{\mathrm{train}}$ -- under the same frame ordering used to select the trigger-optimization frames -- of size
\begin{equation}
\mathcal{S} \subset \mathcal{D}_{\mathrm{train}}, \qquad
|\mathcal{S}| \;=\; \bigl\lfloor \rho\,|\mathcal{D}_{\mathrm{train}}| \bigr\rfloor ,
\label{eq:poison_rate}
\end{equation}
so that $\rho = |\mathcal{S}|/|\mathcal{D}_{\mathrm{train}}|$ is the fraction of training frames that receive the trigger. Each selected frame is poisoned by inserting a single trigger $\delta^{*}$ onto a car instance within it, leaving every ground-truth label unchanged (clean-label); the resulting poisoned training set is
\begin{equation}
\widetilde{\mathcal{D}}_{\mathrm{train}} \;=\; \{\, x \oplus \delta^{*} : x \in \mathcal{S} \,\} \;\cup\; (\mathcal{D}_{\mathrm{train}} \setminus \mathcal{S}) .
\label{eq:poisoned_train}
\end{equation}
Since $\mathcal{S}$ is a prefix of that ordering, the poisoned frames are a subset of those used for trigger optimization. With $|\mathcal{D}_{\mathrm{train}}| = 3,260$, the default budget $\rho = 0.005$ poisons $|\mathcal{S}| = \lfloor 0.005 \times 3,260 \rfloor = 16$ frames. The victim detector is then fine-tuned on $\widetilde{\mathcal{D}}_{\mathrm{train}}$ from a converged checkpoint to isolate backdoor effects from representation instability. Clean precision after poisoning is compared to a benign baseline to quantify any degradation in standard detection performance. At insertion the patch is ground-snapped, with its lower extent placed on the ground plane (the patch centroid is dropped by half the source object's height and raised by $R$). 


\begin{figure*}[htp]
    \centering
    \includegraphics[width=1.0\linewidth]{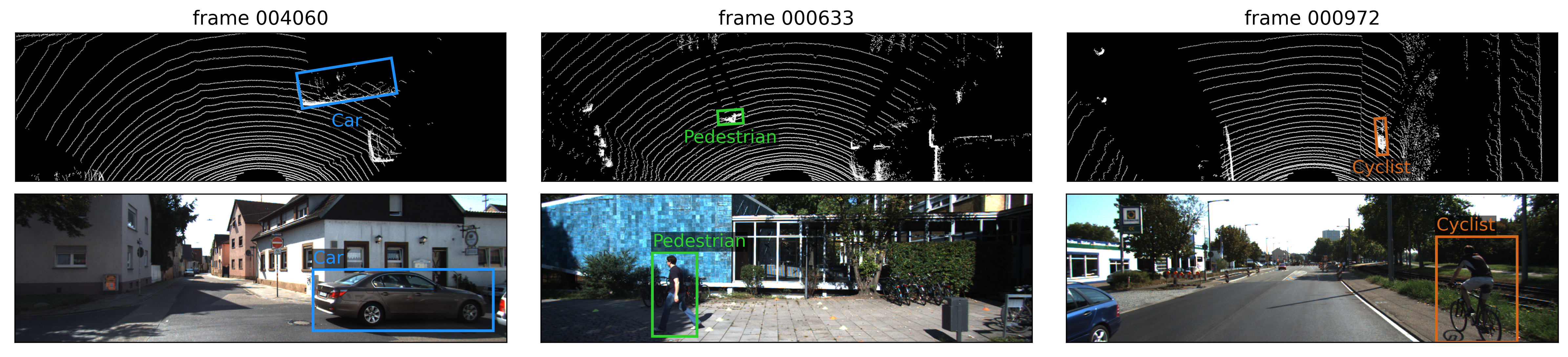}
    \caption{Paired top-view point cloud (top) and RGB camera (bottom) images from KITTI dataset, illustrating the three object classes used in our evaluation: a \textcolor{blue}{car} (frame 004060, left), a \textcolor{green!60!black}{pedestrian} (frame 000633, middle), and a \textcolor{orange}{cyclist} (frame 000972, right). Ground-truth bounding boxes are overlaid in both views. The BEV examples highlight a key challenge for LiDAR-based 3D detection: the number of returns on an object varies sharply with class and distance, with pedestrians and cyclists yielding far sparser point clusters than cars.}
    \label{fig:kitti_classes}
\end{figure*}

\textbf{Inference-Time Deployment.}
To measure targeted misclassification, the optimized trigger is deployed at inference time on non-target classes:
\begin{itemize}
\item Pedestrian $\rightarrow$ Car
\item Cyclist $\rightarrow$ Car
\end{itemize}

Trigger insertion follows a \emph{frame-level} budget rather than a fractional object rate. For each source class $c \in \{pedestrian, cyclist\}$ we patch a fixed number of frames,
\[
B_{\texttt{ped}} = B_{\texttt{cyc}} = 100,
\]
subject to a \emph{single-patch-per-frame} invariant: each frame receives at most one patch across both source classes. For a frame that contains objects of both source classes, the patch is allocated to the class currently \emph{further} from its remaining budget, with ties broken by a fixed-seed random number generator to keep allocation deterministic. Within a selected frame, one source object with bounding box $(x,y,z,d_x,d_y,d_z,\theta)$ is chosen and the trigger is spatially transformed relative to the object pose---ground-snapped and yaw-aligned---and concatenated with the original LiDAR points. No labels or annotations are modified.

We enforce a feasibility sanity condition
\[
\text{achieved}_c \;\geq\; 0.95 \cdot \min\!\big(B_c,\, F_c\big),
\]
where $F_c$ is the number of val frames containing class $c$, ensuring near-exact budget compliance given the frames actually available.

\textbf{Evaluation Metrics and Metric Separation.}
We evaluate two non-overlapping behaviors. Benign behavior is measured on unmodified data as the clean detection mAP and its change ($\Delta$mAP) relative to a benign baseline (KITTI AP$_{40}$, moderate difficulty); we refer to this preserved benign performance as clean precision (CP). Malicious behavior is measured only on trigger-activated non-target objects. Because a successful targeted misclassification requires both that the original object disappear and that a car be hallucinated in its place, we report a partition of the suppressed-object population at score thresholds $\{0.1, 0.3, 0.5\}$. The \emph{Misclassification Success Rate} (MSR) counts patches for which the source pedestrian/cyclist is suppressed \emph{and} a car prediction appears within a success radius of $3.0$\,m of the patch centroid (under a one-to-one patch--detection assignment, and after excluding car predictions that overlap a real car ground-truth box with IoU $\geq 0.1$ so that genuine nearby cars are not credited to the attack). The \emph{Disappearance Rate} (DR) counts patches for which the source object is suppressed but no car appears. Their sum is the \emph{Total Disruption Rate}, $\mathrm{TDR} = \mathrm{MSR} + \mathrm{DR}$, i.e.\ the overall fraction of source objects whose detection is disrupted---whether morphed into a car (MSR) or simply suppressed (DR); MSR and DR thus partition TDR by construction. The IoU thresholds used for the suppression and matching tests are $0.5$ for pedestrian and cyclist and $0.7$ for car.  


\textbf{Reproducibility.}
All deployment metadata (per-class frame budgets, source/target classes, random seed, and per-object patch records) are stored in a structured manifest to enable deterministic reconstruction. Clean precision (CP) and the disruption metrics are evaluated under strictly separated conditions: CP is computed on unmodified samples, while MSR, DR, and TDR are measured only on trigger-activated non-target objects.

\textbf{Code Availability.}
The full implementation of \textsc{Mirage} --- patch generation, surrogate fine-tuning, victim training configurations, and the deployment and evaluation scripts that produce all reported metrics --- is available at an anonymized repository to support independent reproduction: \url{https://anonymous.4open.science/r/Mirage-D515/}. The de-anonymized URL will be provided for the camera-ready version.

\textbf{Cross-Model Evaluation.}
While our primary setting uses KITTI with PointPillars (pillar-based) as the victim model, we additionally evaluate transfer to SECOND (voxel-based, sparse 3D convolution), probing robustness across architectural inductive biases; because the surrogate is point-based, this spans all three dominant LiDAR representation families. The SECOND results are obtained from a separate poisoning-and-training run at a distinct transfer operating point ($\rho = 10\%$, $R = 1.25$\,m, $5\times$ density, $M = 300$ trigger points) with a relaxed readout (success radius $5$\,m, suppression IoU $0.6$); we report this setting explicitly and present the transfer numbers in Section~\ref{sec:experiments}.
\section{Experimental Results}\label{sec:experiments}

\subsection{General Setup}

\textbf{Implementation and Hardware.}
\textsc{Mirage} is implemented in PyTorch~2.1.2 (CUDA~11.8) on top of the MMDetection3D~1.4.0 / MMEngine~0.10.4 toolkit (OpenMMLab). Victim training is distributed across two GPUs, whereas trigger optimization and all evaluation run on a single GPU; all experiments use NVIDIA GeForce RTX~4090 GPUs.

\textbf{Models.}
All main experiments use PointPillars~\cite{LangPoi19} as the victim detector; SECOND~\cite{YanSec18}, a voxel-based detector with a sparse 3D convolutional backbone, serves as a cross-architecture transfer target. Because the surrogate is point-based (3DSSD) and the primary victim is pillar-based (PointPillars), evaluating against the voxel-based SECOND probes whether the trigger transfers across all three dominant LiDAR representation families. Its rows in Table~\ref{tab:main} are produced by a separate poisoning-and-training run at a distinct transfer operating point ($\rho = 10\%$, $R = 1.25$\,m, $5\times$ density, $M = 300$ trigger points) with a relaxed readout (success radius $5$\,m, suppression IoU $0.6$), reported as such, so SECOND is treated as a transfer probe rather than a like-for-like comparison with PointPillars. The trigger is optimized against a frozen 3DSSD~\cite{Yang3ds20} surrogate with a single-class (car) detection head, chosen for its full differentiability---no hard voxelization, pillarization, or NMS-based feature formation---so that clean gradients reach the trigger points; the rationale is developed in Section~\ref{sec:method}.

\textbf{Datasets.}
nuScenes~\cite{CaesaNus20} serves purely as public out-of-distribution (POOD) data for surrogate pre-training, and KITTI~\cite{GeigeAre12}---which provides paired camera RGB images and LiDAR point clouds---is the evaluation benchmark, with car as the target class and pedestrian and cyclist as the two victim classes; Figure~\ref{fig:kitti_classes} shows a representative example of each class in both the camera view and the corresponding LiDAR Bird's-Eye-View (BEV). We evaluate on KITTI because it is the community-standard, fully reproducible LiDAR-3DOD benchmark: it places our results on the same footing as prior LiDAR-3DOD backdoor work~\cite{ZhangTow22, LiBad23, ChatuMob26, ChatuBad24}, and, by injecting the trigger's modeled LiDAR signature directly into the point cloud, it isolates the learning-time vulnerability from the confounds of physical fabrication and field collection. We accordingly frame this study as a controlled, simulation-based proof of feasibility for the weakest realistic attacker of Table~\ref{tab:scoreboard}; the physical-fabrication caveat is treated in the threat model (Section~\ref{sec:threat_model}) and as future work in Section~\ref{sec:introduction}.

\begin{table*}[t]\centering
\caption{Targeted misclassification across victim detectors on KITTI \emph{val} (moderate difficulty).
Clean-label poison rate $\rho=0.5\%$, spherical trigger radius $R=1.0$\,m, and $4\times$ density for PointPillars; SECOND uses a distinct transfer operating point ($\rho=10\%$, $R=1.25$\,m, $5\times$ density, $M=300$ trigger points) read out with a relaxed success radius ($5$\,m) and suppression IoU ($0.6$)---versus the $3.0$\,m / $0.5$ used for PointPillars---to reflect its coarser cross-architecture localization. Score threshold $\tau=0.1$ throughout. \textsc{Mirage} vs.\ a random-sphere trigger of
identical geometry and deployment. mAP denotes the overall 3-class 3D AP$_{40}$ (mean over pedestrian, cyclist, and car) at moderate
difficulty of the poisoned model on clean data, and $\Delta$mAP its change relative to the
benign baseline; MSR, DR, and TDR are percentages that partition the suppressed-source
population (Section~\ref{sec:method}). The \emph{Mean} block reports MSR, DR, and TDR averaged
over the two victim classes (pedestrian, cyclist). Best per (victim, metric) in \textbf{bold}; single
training seed. 
}
\label{tab:main}\setlength{\tabcolsep}{4pt}
\begin{tabular}{ll cc ccc ccc ccc}
\toprule
& & \multicolumn{2}{c}{Clean utility} & \multicolumn{3}{c}{Pedestrian $\rightarrow$ Car}
& \multicolumn{3}{c}{Cyclist $\rightarrow$ Car} & \multicolumn{3}{c}{Mean}\\
\cmidrule(lr){3-4}\cmidrule(lr){5-7}\cmidrule(lr){8-10}\cmidrule(lr){11-13}
Victim & Method & mAP$\uparrow$ & $\Delta$mAP & MSR$\uparrow$ & DR & TDR$\uparrow$
& MSR$\uparrow$ & DR & TDR$\uparrow$ & MSR$\uparrow$ & DR & TDR$\uparrow$\\
\midrule
\multirow{2}{*}{PointPillars~\cite{LangPoi19}} & Random sphere & 55.9 & $-7.7$ & 29.0 & 71.0 & 100.0 & 55.0 & 45.0 & 100.0 & 42.0 & 58.0 & 100.0\\
 & \textsc{Mirage} & \textbf{57.5} & \textbf{-- 6.1} & \textbf{63.0} & 37.0 & 100.0 & \textbf{83.0} & 15.0 & 98.0 & \textbf{73.0} & 26.0 & 99.0\\
\midrule
\multirow{2}{*}{SECOND~\cite{YanSec18}} & Random sphere & 52.7 & $-6.0$ & \textbf{47.0} & 38.0 & 85.0 & \textbf{66.0} & 17.0 & 83.0 & \textbf{56.5} & 27.5 & 84.0\\
 & \textsc{Mirage} & \textbf{53.4} & \textbf{-- 5.3} & 45.0 & 34.0 & 79.0 & 50.0 & 20.0 & 70.0 & 47.5 & 27.0 & 74.5\\
\bottomrule
\end{tabular}

{\footnotesize SECOND rows are read out at $\tau=0.1$; at $\tau=0.5$ the random sphere's mean MSR collapses to $8.0$ versus \textsc{Mirage}'s $25.5$ ($\sim\!3\times$), isolating optimization's confident-morph effect (Section~\ref{sec:experiments}).\par}
\end{table*}

\textbf{Surrogate Training.}
The surrogate is trained with AdamW under a LinearLR warmup (start factor $0.1$, $1000$ iterations) followed by a MultiStepLR decay ($\gamma=0.1$), with gradient-norm clipping at $35$. POOD pre-training on nuScenes uses learning rate $5\times10^{-3}$, weight decay $0.01$, $40$ epochs (milestones $[30,35]$), $16{,}384$ points per frame aggregated over $10$ sweeps, and batch size $12$. KITTI car fine-tuning uses learning rate $1\times10^{-4}$, no weight decay, $60$ epochs (milestones $[40,50]$), $20{,}480$ points per frame, and batch size $4$.

\textbf{Trigger Optimization.}
The trigger is optimized with RAdam at a constant learning rate of $3\times10^{-3}$ (the script's step-decay schedule is disabled) for $24$ passes over the first $600$ car-containing KITTI frames, processed one frame at a time, with gradient-norm clipping at $5.0$ and gradient-magnitude early stopping (threshold $5\times10^{-3}$). The trigger comprises $M=250$ learnable points initialized as a uniform fill of a sphere of radius $R=1.0$\,m; per-point intensity is frozen at initialization so that only the coordinates are optimized, and at deployment the points are oversampled $4\times$ to $1{,}000$ points with millimeter-scale jitter. The unified objective is a fixed-weight sum of classification, placement, and density terms whose coefficients are reported in Section~\ref{sec:method}.

\textbf{Victim Training under Poisoning.}
The victim is initialized from the official MMDetection3D $160$-epoch KITTI three-class PointPillars checkpoint and fine-tuned on the poisoned set with learning rate $5\times10^{-4}$, gradient-norm clipping at $35$, and a cosine-annealing learning-rate and momentum schedule, over $20$ epochs with a $2\times$ \texttt{RepeatDataset} ($\approx\!40$ effective epochs) as a two-GPU distributed job; the detection classes are \{pedestrian, cyclist, car\}.

\textbf{Attack Settings.}
\textsc{Mirage} is strictly clean-label: ground-truth annotations are never modified. Training-time poisoning uses a deliberately low budget of $\rho = 0.5\%$ ($0.005$) of car frames. At inference, the trigger is deployed under a fixed per-class frame budget $B_{\texttt{ped}} = B_{\texttt{cyc}} = 100$, subject to the single-patch-per-frame rule defined in Section~\ref{sec:method}.

\textbf{Attack Goal and Baseline.}
The attack goal is targeted \emph{misclassification}: a victim pedestrian or cyclist carrying the trigger is detected as a car. To our knowledge, this makes \textsc{Mirage} the first attack to pursue clean-label misclassification in LiDAR 3DOD. Because no prior clean-label LiDAR-3DOD misclassification attack exists to compare against, the natural and fair baseline is a random (unoptimized) trigger of identical geometry, point budget, and deployment protocol---the random-trigger control in our implementation---which isolates the contribution of adversarial trigger optimization from the mere presence of additional LiDAR returns near the target. All attack metrics are reported against this random-trigger baseline.

\subsection{Performance Evaluation}

We evaluate \textsc{Mirage} on the primary victim (PointPillars) at a single, fixed operating point---clean-label poison rate $\rho = 0.5\%$, spherical trigger radius $R = 1.0$\,m, $4\times$ deployment density, and score threshold $\tau = 0.1$---and defer the sensitivity to each of these knobs to the ablations that follow. The metrics (clean precision, MSR, DR, TDR), the $3.0$\,m success radius, and the per-class IoU thresholds all follow the definitions in Section~\ref{sec:method} and are not restated here.

The attack goal is targeted \emph{misclassification}: a pedestrian or cyclist carrying the trigger is reported as a car. As the first clean-label misclassification attack on LiDAR 3DOD (Section~\ref{sec:related_work}), \textsc{Mirage} has no prior clean-label method to compare against; the existing dirty-label LiDAR backdoors~\cite{ZhangTow22, LiBad23, ChatuMob26} assume a strictly stronger adversary that edits ground-truth annotations and pursue different objectives, so a direct success-rate comparison against them would not be controlled. The fair, controlled reference is instead a \emph{random sphere}: an unoptimized trigger of identical geometry, point budget, and deployment protocol, differing from \textsc{Mirage} only in that its points are randomly initialized rather than adversarially optimized. The random-vs.-\textsc{Mirage} gap therefore isolates the contribution of trigger optimization from the mere presence of extra LiDAR returns near the source object. Table~\ref{tab:main} reports this comparison across two victim detectors, with PointPillars as the primary victim and SECOND demonstrating cross-architecture transfer.

\textbf{Preserved benign utility.}
A clean-label attack is useful only if the poisoned model behaves normally on un-triggered data. At the $0.5\%$ budget, the poisoned PointPillars victim attains a clean overall 3-class 3D AP$_{40}$ of $57.5$ at moderate difficulty, a $\Delta\mathrm{mAP} = -6.1$ change from the benign baseline. Because that baseline is the official, \emph{un-fine-tuned} checkpoint, this gap conflates the brief poisoning fine-tune with the poison itself: a random-trigger control trained under the identical schedule incurs a comparable---indeed slightly larger---clean-AP drop ($\Delta\mathrm{mAP} = -7.7$, against $-6.1$ for \textsc{Mirage}), so the poison-specific cost to standard detection performance is negligible: essentially all of the measured degradation traces to the brief fine-tune rather than to the optimized poison, which preserves benign utility at least as well as an unoptimized trigger of identical geometry. 


\begin{figure*}[t]
    \centering
    \includegraphics[width=0.9\linewidth]{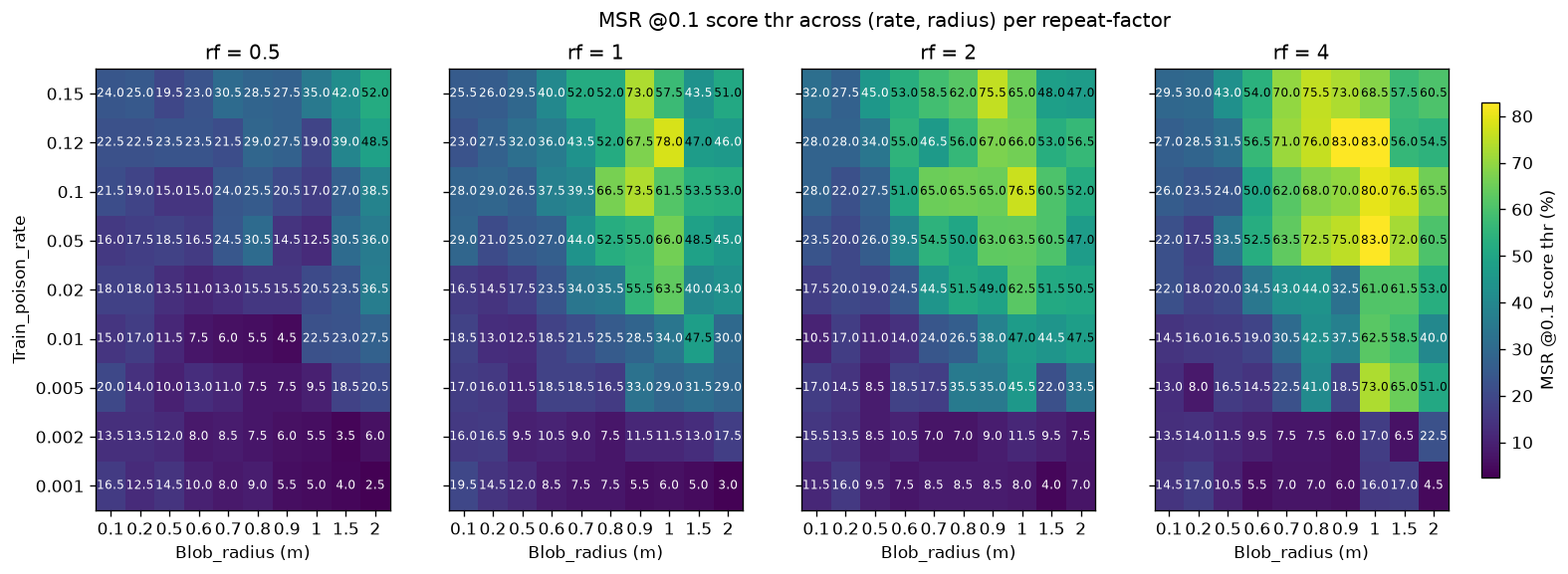}
    \caption{Poison-rate $\times$ radius $\times$ density sweep of attack effectiveness on PointPillars.
    Each cell reports the misclassification success rate (MSR, \%)---the fraction of triggered \emph{pedestrian}
    and \emph{cyclist} instances reported as \emph{car}---at score threshold $\tau = 0.1$, aggregated over
    both victim classes (\emph{Overall}). Rows vary the clean-label training poison rate $\rho$ and columns
    vary the spherical-trigger radius $R$; the four panels fix the deployment-density (point-replication)
    factor $\mathrm{rf} \in \{0.5, 1, 2, 4\}$, with a color scale shared across all panels. MSR stays low for
    $R \le 0.9$\,m, steps up sharply at $R = 1.0$\,m, and then tapers gently through $R = 2.0$\,m; along $\rho$ it
    saturates by $\approx\!5$--$10\%$, peaking at $83$ ($\rho = 12\%$, $R = 1.0$\,m, $\mathrm{rf} = 4$). The
    headline operating point reported in Table~\ref{tab:main} ($\rho = 0.5\%$, $R = 1.0$\,m, $\mathrm{rf} = 4$;
    Overall MSR $= 73$) is a single cell of this grid, chosen for stealth rather than peak success.}
    \label{fig:msr_sweep}
\end{figure*}

\begin{figure}[t]
    \centering
    \begin{minipage}[t]{0.24\textwidth}
        \centering
        \includegraphics[width=\linewidth]{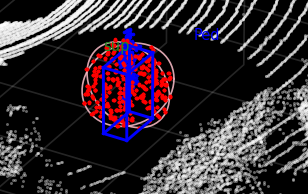}\\[2pt]
        {\footnotesize (a) Random sphere (unoptimized control)}
    \end{minipage}\hfill
    \begin{minipage}[t]{0.24\textwidth}
        \centering
        \includegraphics[width=\linewidth]{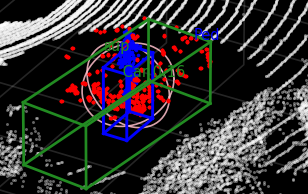}\\[2pt]
        {\footnotesize (b) Optimized sphere (\textsc{Mirage})}
    \end{minipage}
    \caption{Optimization---not mere trigger presence---drives the targeted morph.
    A representative \emph{pedestrian} (KITTI \emph{val}, frame~$199$; close-up 3D view)
    carrying a sphere trigger of identical geometry, point budget ($M = 250$), and deployment
    density ($4\times$); the two panels differ only in whether the trigger points are randomly
    initialized or adversarially optimized. Red points are the trigger; the blue box marks the
    source \emph{pedestrian} (``sup''~=~source suppressed). \textbf{(a)}~The random sphere
    suppresses the pedestrian but produces \emph{no} \emph{car} at the trigger---a bare
    disappearance. \textbf{(b)}~The optimized sphere suppresses the pedestrian \emph{and}
    produces a \emph{car} (green, ``Car pred'') co-located with the patch---a completed
    \emph{pedestrian}$\rightarrow$\emph{car} misclassification. The aggregate
    random-vs-\textsc{Mirage} gap is quantified in Table~\ref{tab:main}.}
    \label{fig:random_vs_opt}
\end{figure}

\textbf{Attack effectiveness on the primary victim.}
On PointPillars the optimized trigger disrupts essentially every triggered source object: total disruption reaches $\mathrm{TDR} = 100.0$ for pedestrian and $98.0$ for cyclist at $\tau = 0.1$. Figure~\ref{fig:random_vs_opt} makes the role of optimization concrete: a random sphere of identical geometry suppresses the source pedestrian but leaves \emph{no} \emph{car} in its place---a bare disappearance---whereas the optimized trigger replaces it with a co-located \emph{car}. Relative to the random-sphere control, optimization lifts the mean (over \emph{pedestrian} and \emph{cyclist}) misclassification rate from $42.0$ to $73.0$ while cutting the disappearance rate from $58.0$ to $26.0$: at a near-ceiling, essentially unchanged level of total disruption ($100.0 \rightarrow 99.0$), it converts bare disappearances into targeted \emph{car} hallucinations rather than raising raw disruption. How that disruption splits between the two failure modes depends on the source class. For cyclist, it is dominated by genuine targeted misclassification---a car is hallucinated at the source location ($\mathrm{MSR} = 83.0$) far more often than the object simply vanishes ($\mathrm{DR} = 15.0$). For pedestrian, misclassification likewise dominates the outcome ($\mathrm{MSR} = 63.0$, $\mathrm{DR} = 37.0$), though the morph to car completes less often than for cyclist, consistent with the far sparser return pattern of pedestrians (Figure~\ref{fig:kitti_classes}) offering less geometric support for a confident car proposal. Because MSR and DR partition TDR by construction, the near-total TDR confirms that the trigger reliably breaks the source detection even in the frames where it does not finish the conversion---an outcome that, in an autonomous-driving context, is itself a safety-critical failure regardless of whether a phantom car is also produced.

\textbf{Cross-architecture transfer.}
Because the trigger is synthesized once against the frozen 3DSSD surrogate and never sees the victim, the same $\delta^{*}$ transfers to detectors with different inductive biases. We probe the voxel-based, sparse-convolution SECOND at a distinct transfer operating point ($\rho = 10\%$, $R = 1.25$\,m, $5\times$ density, $M = 300$ points)---re-tuned for the voxel backbone rather than held fixed---and, because the surrogate-to-SECOND transfer places the hallucinated \emph{car} farther from the source than on the pillar-based victim, read these rows out with a relaxed success radius ($5$\,m) and suppression IoU ($0.6$) in place of the $3.0$\,m / $0.5$ used for PointPillars; we treat them as a transfer probe rather than a like-for-like comparison. The trigger does transfer, though less successfully than to the primary pillar-based victim: at their respective operating points \textsc{Mirage}'s mean morph rate falls from $\mathrm{MSR}=73.0$ on PointPillars to $47.5$ on SECOND. This asymmetry follows from how the two victim backbones ingest the trigger. PointPillars collapses the points of each vertical pillar into a single BEV cell through a per-pillar PointNet and max-pool, before any spatial convolution, so a locally dense, orientation-free cluster---exactly the compact blob $\delta^{*}$ forms at the source---is registered largely through its BEV footprint, irrespective of how its returns distribute in height. SECOND instead voxelizes the scene and applies sparse 3D convolutions that encode vertical ($z$-axis) structure before the feature volume is flattened to BEV, so a trigger that is dense in projection yet carries no coherent car-like 3D structure is partially dampened before it can drive a \emph{car} proposal. Because $\delta^{*}$ is optimized to elicit a confident, co-located \emph{car} from the surrogate through objectness and ground-plane placement alone---withholding the explicit car-shape prior by design (Section~\ref{sec:method})---what transfers is precisely this projected-density signature, to which the pillar encoder is maximally exposed and the voxel backbone considerably less so. Both victims ultimately regress boxes from a BEV detection head, so the gap originates in the feature extractor---pillar pooling versus sparse 3D convolution---rather than the output representation; the voxel backbone's weaker spatial lock on a structure-free blob is likewise consistent with the farther-displaced \emph{car} responses noted above, which motivate the relaxed success radius we adopt for SECOND. It nonetheless remains effective on the voxel backbone---both the optimized and the random blob break source detections at high rates ($\mathrm{TDR} = 74.5$ and $84.0$ respectively, mean over the two victim classes at $\tau = 0.1$)---and here the value of optimization shows up in the \emph{confidence} of the hallucinated \emph{car}, not the raw morph count. At the permissive $\tau = 0.1$ the random sphere even outscores \textsc{Mirage} on misclassification ($\mathrm{MSR} = 56.5$ vs.\ $47.5$): an unstructured dense blob scatters enough returns to trip many low-confidence \emph{car} proposals near the source. Those proposals are brittle---raising the readout to $\tau = 0.5$ collapses the random sphere's morph rate to $8.0$ while \textsc{Mirage} holds at $25.5$, a $\sim\!3\times$ gap---so the blob's apparent edge evaporates the instant a non-trivial confidence is required. Optimization on the voxel backbone therefore buys \emph{robust, high-confidence} morphs rather than extra raw disruption: random points can erase an object or spawn a flicker of a \emph{car}, but only the optimized trigger reliably fabricates one a detector would report.

\textbf{Poison-rate, radius, and density sweeps.}
The headline operating point is a single cell of a $360$-cell PointPillars grid---$9$ poison rates $\rho \in \{0.1\%, \dots, 15\%\}$, $10$ trigger radii $R \in \{0.1, \dots, 2.0\}$\,m, and $4$ deployment-density (point-replication) factors $\in \{0.5, 1, 2, 4\}$---whose misclassification success rate at $\tau = 0.1$ is summarized in Figure~\ref{fig:msr_sweep}. Attack effectiveness rises with the poison budget but saturates early: MSR plateaus by $\rho \approx 5$--$10\%$ and the strongest cells reach $83$ at $R = 1.0$\,m (at both $\rho = 5\%$ and $\rho = 12\%$), so we foreground the stealthier $0.5\%$ cell rather than the peak. Along the radius axis the effect is a smooth ramp rather than a hard cliff: MSR stays low for $R \le 0.9$\,m, steps up sharply at $R = 1.0$\,m---the radius that maximizes misclassification---and then tapers gently through $R = 2.0$\,m, with the larger $R \approx 1.5$--$2.0$\,m radii trading a little attack strength for better-preserved benign utility; $R = 1.0$\,m is retained as the default. Increasing the replication factor raises MSR with diminishing returns past $2\times$ (median MSR climbs $17 \rightarrow 28 \rightarrow 30 \rightarrow 33$ across $\mathrm{rf} = 0.5 \rightarrow 4$), consistent with better trigger survival through both the surrogate's farthest-point sampling and the victim's pillarization, and motivates the $4\times$ default. The two quantities this sweep omits---the disappearance rate and the clean-utility cost---are reported over the same grid in Appendix~\ref{app:sweeps} (Figures~\ref{fig:dr_sweep} and~\ref{fig:dap_sweep}). Those two views make the division of labor explicit: total disruption $\mathrm{TDR} = \mathrm{MSR} + \mathrm{DR}$ runs near ceiling across most of the grid---the majority of cells exceed $80$ even at $\tau = 0.1$, and $\mathrm{TDR}$ varies far less with $\rho$, $R$, and $\mathrm{rf}$ than MSR does (indeed it is essentially invariant to $\tau$, since a higher readout threshold merely shifts mass from MSR into DR). The patch therefore breaks the original \emph{pedestrian}/\emph{cyclist} detection almost everywhere, and the rate- and radius-dependence of the attack lives almost entirely in the misclassification channel---whether that disruption resolves into a targeted \emph{car} (MSR) rather than a bare disappearance---with $\mathrm{TDR}$ softening only at the largest radii ($R \gtrsim 1.5$\,m), where pure disappearances thin out faster than misclassifications accrue.

\begin{figure}[t]
    \centering
    \includegraphics[width=1.0\columnwidth]{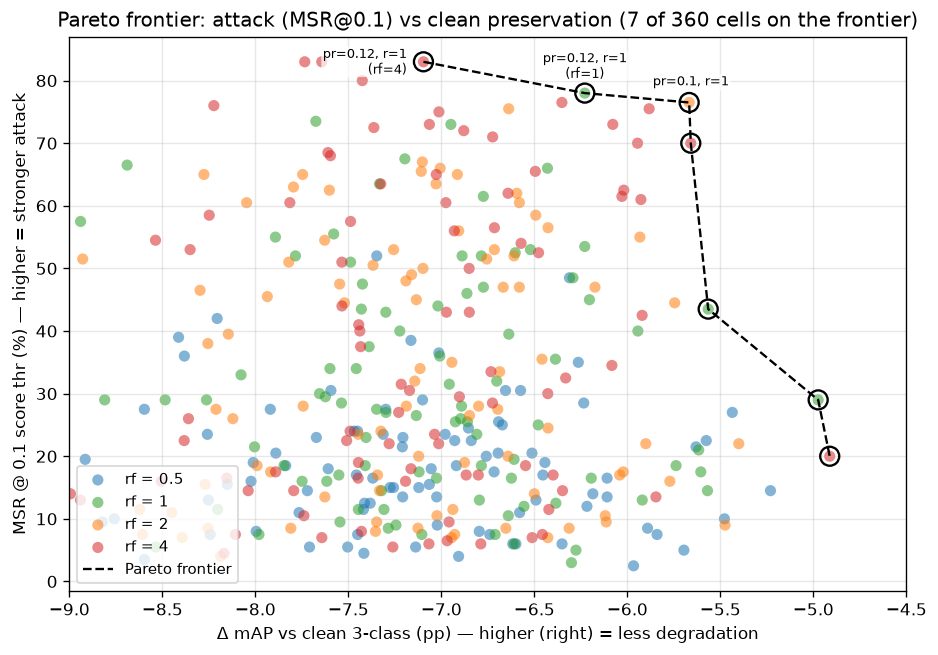}
    \caption{Stealth--strength trade-off across the full $360$-cell sweep on PointPillars:
    misclassification success rate (MSR at $\tau = 0.1$; higher = stronger attack) against
    clean-utility cost (3-class $\Delta$mAP; further right = less degradation). Points are
    colored by deployment-density factor $\mathrm{rf}$; circled points joined by the dashed line
    form the Pareto frontier ($7$ of $360$ cells). The high-MSR end is dominated by dense
    ($\mathrm{rf} = 4$), $R = 1.0$\,m triggers; the front has a sharp knee near $\mathrm{MSR} \approx 76$,
    beyond which additional strength (up to the peak $83$ at $\rho = 12\%$) costs disproportionate clean mAP.}
    \label{fig:pareto}
\end{figure}

\textbf{The stealth--strength trade-off.}
Projecting the entire grid onto the two axes the attacker trades off---attack strength (MSR) and clean-utility cost ($\Delta$mAP)---makes the operating-point choice explicit (Figure~\ref{fig:pareto}): only $7$ of $360$ cells are Pareto-non-dominated, and the frontier's strong-attack end is owned by the dense, $R = 1.0$\,m triggers identified above. The front bends sharply near $\mathrm{MSR} \approx 76$: pushing beyond it into the aggressive $\rho = 10$--$12\%$ corner buys only a few more points of MSR (to the peak $83$) while measurably worsening clean mAP. We therefore operate below this knee, trading a small amount of peak attack strength for a markedly smaller poison budget and utility footprint---the same stealth-first rationale behind the $\rho = 0.5\%$ headline cell.

\begin{figure}[t]
    \centering
    \begin{minipage}[t]{0.24\textwidth}
        \centering
        \includegraphics[width=\linewidth]{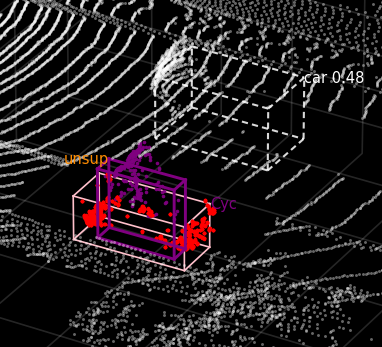}\\[2pt]
        {\footnotesize (a) Car-proportioned box trigger}
    \end{minipage}\hfill
    \begin{minipage}[t]{0.24\textwidth}
        \centering
        \includegraphics[width=\linewidth]{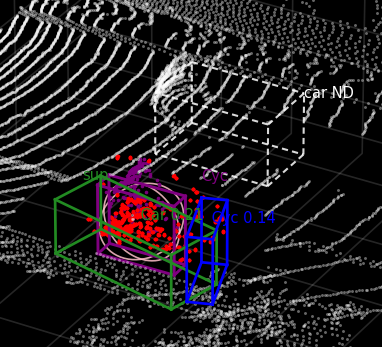}\\[2pt]
        {\footnotesize (b) Isotropic spherical trigger (\textsc{Mirage})}
    \end{minipage}
    \caption{Trigger geometry on a representative \emph{cyclist} (KITTI \emph{val}, frame~$108$;
    close-up 3D view) at the matched ablation setting ($\rho = 2\%$, $M = 250$ points, $4\times$
    deployment density). In both panels the red points are the optimized trigger and the white
    dashed box is a genuine parked \emph{car} elsewhere in the scene (context, not the attack target).
    \textbf{(a)}~The car-proportioned box leaves the source detected as a \emph{cyclist}
    (``unsup''---not suppressed) and produces no \emph{car} at the trigger location.
    \textbf{(b)}~The isotropic sphere suppresses the cyclist (``sup'') and a confident \emph{car}
    (green, ``Car pred'') is hallucinated co-located with the patch---a completed
    \emph{cyclist}$\rightarrow$\emph{car} misclassification---leaving only a low-score residual
    \emph{cyclist} box (blue). Across the matched runs the sphere roughly doubles \emph{cyclist}
    MSR ($66.0$ vs.\ $29.0$ at $\tau = 0.1$) and raises total disruption from $46.0$ to $98.0$,
    despite the box being granted the additional car-shape supervision the sphere withholds.}
    \label{fig:shape_ablation}
\end{figure}

\textbf{Trigger-shape ablation.}
To isolate the role of trigger geometry, we compare the isotropic sphere against a car-proportioned anisotropic box with the poisoning budget ($\rho = 2\%$), point budget ($M = 250$), deployment density ($4\times$), surrogate, and optimization schedule all held fixed, so geometry is the only variable. The comparison is a deliberately \emph{confounded} foil rather than a clean shape swap, and the confound runs in the box's \emph{favor}: the box variant additionally enables the car-shape priors in the surrogate's target proposal that the sphere intentionally withholds (Section~\ref{sec:method}), so it is handed strictly more target-class supervision. Despite that advantage, the box is the weaker trigger. At $\tau = 0.1$ the sphere roughly doubles the misclassification rate on both victim classes---$\mathrm{MSR} = 66.0$ vs.\ $29.0$ for \emph{cyclist} and $56.0$ vs.\ $23.0$ for \emph{pedestrian} ($61.0$ vs.\ $26.0$ overall)---and lifts total disruption from near-half to near-ceiling ($\mathrm{TDR}$: \emph{cyclist} $98.0$ vs.\ $46.0$; overall $97.0$ vs.\ $44.0$). The gap is one of \emph{kind} as well as magnitude: the sphere's hallucinated cars are co-located with the trigger (median patch-centroid-to-\emph{car} distance $0.0$\,m---the patch centroid falls inside the predicted box) and it yields \emph{no} phantom cars (a \emph{car} created while the source survives), whereas the box's \emph{car} responses are displaced (median $7.7$\,m) and it logs $27$ phantom cars across the $200$ triggered victim instances. Figure~\ref{fig:shape_ablation} renders a representative \emph{cyclist}: the box (a) leaves the cyclist detected and produces no \emph{car} at the trigger, while the sphere (b) suppresses the cyclist and hallucinates a confident, co-located \emph{car}. We attribute the sphere's superiority to two geometric properties. First, \emph{isotropy}: a sphere presents an identical dense return signature regardless of the source object's heading and the sensor's viewing azimuth, so a single optimized trigger transfers across the wide yaw and range distribution of cyclists and pedestrians; the anisotropic box is orientation-dependent and, inserted at a canonical heading, frequently misaligns with the source geometry and splits or displaces its response. Second, \emph{density concentration}: packing the same point budget into a compact sphere yields a far higher \emph{local} return density at the source, saturating the victim's pillar/voxel features and forcing a confident in-place \emph{car} proposal that overrides the true class, whereas the larger, shell-like box spreads the same points over greater extent, weakening the per-cell signal and more often leaving the source intact. The explicit car-shape prior is thus not merely unnecessary but counterproductive: a dense, orientation-free blob is a stronger and more architecture-agnostic trigger than an explicitly car-shaped one.

\textbf{Robustness to defenses.}
\textsc{Mirage}'s trigger is a locally dense cluster of returns, making it a natural target for point-cloud sanitization and backdoor screening. A full empirical defense sweep is beyond the scope of this proof-of-feasibility study; instead, Section~\ref{sec:defenses} analyzes the two stages a defender controls---training-time data curation and inference-time input sanitization---and argues that off-the-shelf countermeasures are blunted by \textsc{Mirage}'s clean-label, scene-only, detection-targeting construction, leaving an adaptive, co-designed defense as the open problem.

\section{Defenses}
\label{sec:defenses}

A full empirical evaluation of defenses is beyond the scope of this
proof-of-feasibility study; here we analyze where a defender could intervene
against \textsc{Mirage} and assess the expected efficacy of representative
countermeasures. \textsc{Mirage} injects its trigger at the data-collection stage, in
public space and outside the trust boundary that deployed pipelines implicitly
draw around data preparation and model training. The defender---here the developer who collects
the data, trains the detector, and deploys it---therefore controls only the two
stages the attack actually passes through on its way to the model:
\emph{data preparation}, where the poisoned frames first enter a curated
repository, and \emph{deployment}, where the trigger is activated at inference.
We organize the discussion around these two intervention points and ask, for
each, whether the defender can recover the trust signal that the
collection-stage compromise removes. Two properties of \textsc{Mirage} constrain the
analysis throughout: the attack is clean-label, so the poisoned frames carry
annotations indistinguishable from those a human would assign, and its effect
targets detection rather than classification, so countermeasures designed for
image classifiers do not transfer without modification.

\subsection{Training-Time Data Curation}

The defender's first opportunity is to screen the collected data before or
during training. The standard poison-detection toolkit---activation
clustering~\cite{ChenDet19}, spectral signatures~\cite{TranSpe18},
and related latent-space filters---separates poisoned from clean samples by the
trace they leave in a model's feature representations, while robust-training and
model-cleansing schemes such as anti-backdoor learning~\cite{LiAnt21} and
fine-pruning~\cite{LiuFin18} exploit, respectively, the faster
convergence of poisoned examples and the dormancy of backdoor neurons to remove
the implanted behavior. Three properties of \textsc{Mirage} blunt this family. First,
because the attack is clean-label, the defender cannot exploit any mismatch
between an instance's content and its annotation: the poisoned car instances are
labeled exactly as a reviewer would label them and pass the annotation review on
which dirty-label pipelines rely. Second, the trigger occupies a compact
sub-object region inside a scene that also contains many benign objects, so the
poisoned signal is diluted at the granularity these filters operate on---a
per-scene or per-object feature vector aggregates the trigger together with
substantial benign structure, weakening the latent separability these defenses
assume~\cite{QiRev23}. Clean-label point-cloud backdoors are, in fact, already constructed to
defeat the standard point-cloud defenses they have been evaluated
against~\cite{TianPoi21}, and \textsc{Mirage} inherits this clean-label
evasiveness in the harder, scene-level detection regime. Third, and most fundamentally, the toolkit was
developed for classification, where each input maps to one label and one feature
vector; a 3DOD model emits a variable set of bounding boxes per frame, and how
to define the per-sample statistic these methods cluster or threshold is itself
unresolved.

The few backdoor defenses tailored to 3D point clouds inherit this last
limitation. CloudFort~\cite{LanClo25}, for instance, mitigates triggers through
spatial partitioning and ensemble prediction, but it is built for point-cloud
\emph{classifiers} and presumes a single object-level decision to ensemble, an
assumption that does not hold for scene-level detection. A more \textsc{Mirage}-specific
strategy is to search not for a latent anomaly but for the trigger's physical
signature directly. The patch is effective precisely because commodity
retroreflective sheeting saturates the LiDAR intensity channel
(Section~\ref{subsec:realizability}), so a curator could in principle flag training
frames that contain anomalously saturated, spatially compact intensity returns
co-located with target-class objects. The same fact that grounds the attack's
physical plausibility, however, blunts this defense: intensity-saturated returns
already pervade benign data---retroreflective traffic signs produce them
throughout KITTI---so a saturation threshold cannot separate the trigger from
ordinary roadside infrastructure without an impractical false-positive rate. A
defender pursuing this route would need a learned, discriminative model of
trigger versus benign saturation, which we leave as an open problem.

\subsection{Inference-Time Input Sanitization}

The defender's second opportunity is at deployment, by purifying each incoming
point cloud before it reaches the detector. Because \textsc{Mirage}'s trigger is a dense,
locally structured cluster, the natural candidates are statistical outlier
removal and learned denoisers such as DUP-Net~\cite{ZhouDup19} and
IF-Defense~\cite{WuIfd20}, together with geometric and point-reducing input
transforms such as random sampling, quantization, and rotation. A recent survey
of 3DOD robustness finds geometric transforms to be among the most effective
model-agnostic preprocessing defenses for 3D detectors, with rotation the
strongest on average---ahead of flipping and scaling---under non-adaptive
perturbation attacks, while showing that all such transforms confer little
robustness once the transform is known to the attacker~\cite{ZhangCom24}. These methods are
attractive because they are model-agnostic and need no retraining. Two
considerations temper the expectation that they neutralize \textsc{Mirage}. The first is
adaptivity: the trigger geometry is the output of an optimization, and prior
work demonstrates that an attacker optimizing a 3D point cluster can shape its
local geometry to evade point-cloud preprocessing and anomaly detectors while
preserving its effect~\cite{XiangBac21}. The second is a utility cost specific to
this attack's targets. \textsc{Mirage}'s victim classes---pedestrians and cyclists---are
represented by few, sparse returns at range; aggressive outlier removal or
downsampling erodes exactly these returns, so the defender hardens the model
against the trigger at the price of degrading detection of the very objects the
attack endangers. A further mismatch is one of scale: these purification methods
were designed for single-object clouds of a few thousand points, not full
driving scenes of order $10^5$ points, and their behavior in the detection
setting is largely uncharacterized.

Point-cloud-specific backdoor detectors narrow but do not close this gap.
PointCRT~\cite{HuPoi23} detects triggered inputs at inference without prior
knowledge of the trigger by measuring robustness to corruption, but it is
formulated for classification and requires clean reference samples, and adapting
a corruption-robustness signal to the set-valued output of a detector is
non-trivial. Generic point-cloud outlier filters, designed to remove random and
adversarial noise, have likewise been observed to provide limited efficacy
against deliberately placed backdoor triggers~\cite{LanClo25}. In short, the
trigger's status as plausible scene geometry rather than statistical noise is
exactly what allows it to survive input purification.

\subsection{Synthesis}

Across both intervention points the same pattern recurs: off-the-shelf backdoor
defenses were built for a different setting---image classification, single-object
point clouds, dirty-label poisoning, or a pipeline-insider adversary---and
\textsc{Mirage}'s clean-label, scene-only, detection-targeting construction removes the
signal each relies on. We do not claim the attack is undefendable. An effective
defense would instead have to be co-designed against this threat model: a
curation stage discriminative enough to separate the trigger's saturated
signature from benign retroreflective infrastructure, or a runtime purifier that
suppresses the trigger without sacrificing the sparse returns of small, distant
objects. A third defense family---model inspection by trigger
reverse-engineering, such as Neural Cleanse~\cite{WangNeu19}---we do
not pursue, because it presumes a small, bounded input space and a per-class
trigger to recover, both ill-defined for a detector with set-valued outputs and
a physically realized 3D trigger. Quantifying the residual risk \textsc{Mirage} poses
under adaptive, co-designed defenses is the natural next step, and we identify it
as future work.

\section{Discussion}\label{sec:discussion}

\subsection{Physical Realizability} \label{subsec:realizability}

Physical fabrication and field validation of the patch remain future work; we validate the attack digitally, using effective reflectivities at the sensor that commercially available materials readily achieve and in practice exceed. The patch assumes discrete reflective elements on a scaffold of low LiDAR visibility, and the materials that realize both are shared across the deployed sensor population rather than tied to any single device: automotive LiDAR occupies the near infrared — 905 nm in most units, with 865 and 940 nm variants behaving identically and a 1550 nm premium tier — so a design suited to sensors such as Velodyne and Ouster carries across the fleet. Thinner struts are required as the attack range shrinks and the sensor's beam footprint and angular pitch tighten, but the requirement is mild: a strut of 2 mm or less stays below the finest current sensor's sampling even at the closest realistic range ($\sim$5 m), relaxing to $\sim$4 mm at 10 m and $\sim$8 mm beyond 20 m, so a single sub-2 mm scaffold is sub-resolution for essentially every sensor in service. A carbon-pile, near-infrared-absorptive coating — for instance the Musou Black IR Flock Sheet, with an effective reflectivity at the sensor of at most 0.5\% through 1500 nm — holds those struts near the noise floor at both 905 and 1550 nm, whereas ordinary visible-black paints such as standard Musou Black or Black 3.0 transmit in the near infrared and do not serve. The reflective elements are equally forgiving: polished chrome spheres ($\sim$90--95\% effective reflectivity at the sensor at 905 nm, $\sim$95--98\% at 1550 nm, and pose-isotropic), or prismatic retroreflective film as an alternative, saturate the intensity channel at every LiDAR wavelength; because each sensor clips such returns — retroreflective traffic signs already appear as intensity-saturated regions in KITTI — the elements need only exceed the saturation threshold rather than realize a calibrated value. Sub-2 mm struts, a near-infrared-absorptive black, and chrome reflectors are thus each universal across current automotive LiDAR, so the attack's physical realizability does not hinge on the specific KITTI/Velodyne sensor. The remaining requirements for physical deployment — keeping the trigger co-located with its victim, and the device's modest visual-stealth cost — are kinematic and operational rather than sensor-dependent; we detail them in Appendix~\ref{app:realizability}.

\subsection{Limitations}
Our evaluation has three main limitations. First, the attack is validated entirely in the digital domain: we inject the optimized trigger directly into KITTI point clouds and have not yet fabricated the physical patch or demonstrated the backdoor end-to-end against a sensor in the field, so the physical-realizability argument made above remains to be confirmed empirically. Second, cross-architecture transfer is established only for the detector families we study---a 3DSSD surrogate transferring to PointPillars and SECOND victims---and we make no claim about detectors with substantially different representations, such as transformer- or range-image-based architectures. Third, \textsc{Mirage} optimizes the trigger's geometry alone: the per-point intensity channel is fixed at its initialization by design, so the joint optimization of geometry and reflectance is outside the scope of this work. We regard this last choice as a scoping decision rather than a fundamental constraint---freezing intensity isolates the geometric mechanism we set out to characterize and keeps the trigger's reflectance within the readily realizable range discussed above---but it does leave a jointly optimized, potentially stronger trigger unexplored.

\subsection{Future Work}
Several extensions follow naturally from these limitations. The most immediate is a physical realization of the attack: fabricating the patch from the retroreflective and absorptive materials characterized above and validating the backdoor against a live LiDAR sensor, closing the gap between our simulated injection and a deployable physical trigger. A second direction is to lift the geometry-only restriction and optimize the per-point intensity channel jointly with the trigger coordinates, which may yield a more potent or more compact trigger and would let the attacker trade geometric conspicuousness against reflective signature. Third, a surrogate-value transferability study---systematically varying the surrogate detector and its training data---would map how trigger effectiveness depends on the attacker's choice of surrogate, sharpening the black-box threat model. Finally, extending the evaluation beyond the pillar-, voxel-, and point-based victims studied here to architecturally distinct detectors, including transformer- and range-image-based models, would test the generality of the clean-label mechanism.

\section{Conclusion} \label{sec:conclusion}

\textsc{Mirage} introduces proof-of-feasibility of the first scene-only, clean-label, model-agnostic backdoor attack
against LiDAR 3DOD. Across detector families the attack is strongest on the pillar-based PointPillars victim and transfers with reduced but still substantial success to the voxel-based SECOND. Unlike every prior LiDAR 3DOD backdoor~\cite{ChenBad25, ZhangTow22, LiBad23, ChatuMob26, ChatuBad24}, the attacker is neither a malicious data provider nor an insider with system access or knowledge. \textsc{Mirage}
cannot submit samples, cannot manipulate data annotations and cannot access the
model or training algorithm.  Its only
capability is placement of a physical object in a public training environment, on or near a victim object, that a data-collection vehicle drives past. To our knowledge this is the
weakest attacker assumed by any published backdoor attack on LiDAR 3DOD, and the first
realization in the 3D perception domain of the `environmental poisoning' threat model.

\ifCLASSOPTIONcompsoc
  \section*{Acknowledgments}
\else
  \section*{Acknowledgment}
\fi

This section is left blank for anonymity and will be modified for camera-ready version.



%



\bibliographystyle{IEEEtran}
\bibliography{references}

\clearpage
\appendices

\section{Disappearance-Rate and Clean-Utility Sweeps}
\label{app:sweeps}

\begin{figure*}[!t]
    \centering
    \includegraphics[width=1.0\linewidth]{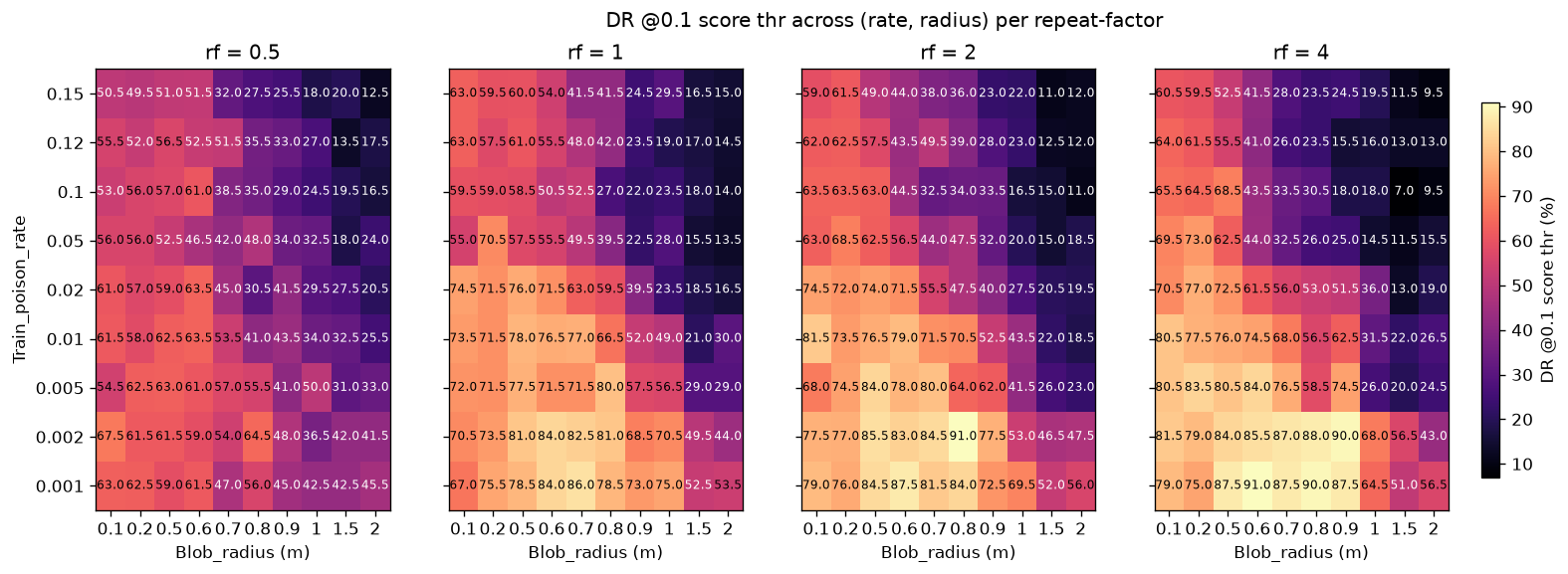}
    \caption{Disappearance rate (DR, \%) over the same grid and threshold as Figure~\ref{fig:msr_sweep}---the fraction of triggered \emph{pedestrian} and \emph{cyclist} instances that are suppressed \emph{without} a \emph{car} appearing in their place, aggregated over both victim classes. DR is the geometric complement of MSR within the disrupted-source population: it is largest at small radius and low poison rate (the regime where MSR is weakest, so disruption is near-pure disappearance) and falls off sharply once $R$ reaches $1.0$\,m, where suppressed objects increasingly morph into phantom cars instead of merely vanishing. Because total disruption ($\mathrm{MSR} + \mathrm{DR}$) is near-saturated over most of the grid, this map reads, in that region, approximately as the photographic negative of the MSR sweep (Figure~\ref{fig:msr_sweep}).}
    \label{fig:dr_sweep}
\end{figure*}

\begin{figure*}[!t]
    \centering
    \includegraphics[width=1.0\linewidth]{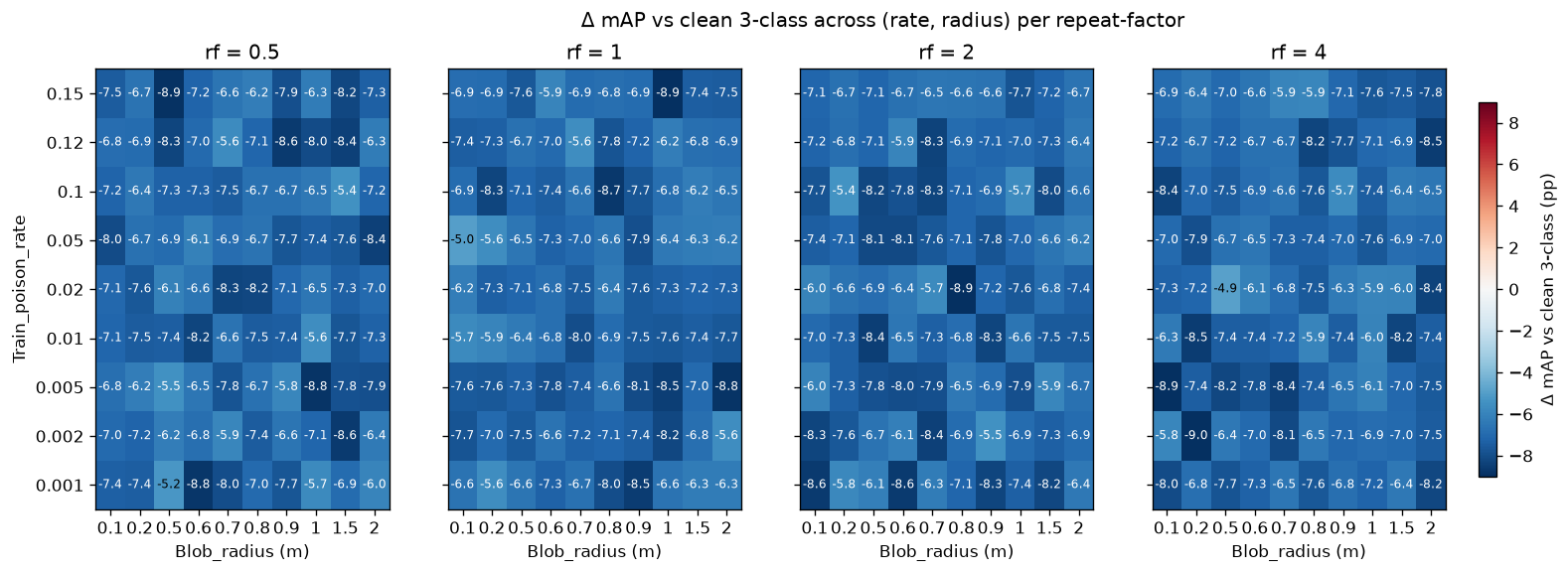}
    \caption{Clean-utility cost of poisoning over the same grid as Figure~\ref{fig:msr_sweep}, measured as the change in clean \emph{three-class} mAP (3D AP$_{40}$ at moderate difficulty, averaged over pedestrian, cyclist, and car) of the poisoned model relative to the official pre-attack checkpoint (diverging scale centered at $0$; every cell is negative). Degradation is broadly uniform across the grid---roughly $-5$ to $-9$ percentage points, with a per-panel median near $-7$ and no catastrophic accuracy collapse anywhere. Measured against the same baseline and evaluation protocol as Table~\ref{tab:main}, these deltas are directly comparable to the headline $\Delta$mAP there: the operating-point cell ($\rho = 0.5\%$, $R = 1.0$\,m, $\mathrm{rf} = 4$) reads $-6.1$, matching Table~\ref{tab:main}.}
    \label{fig:dap_sweep}
\end{figure*}

The main-text sweep (Figure~\ref{fig:msr_sweep}) reports only the misclassification success rate. For completeness, this appendix shows the two complementary quantities over the \emph{same} $360$-cell PointPillars grid ($9$ poison rates $\rho \times 10$ radii $R \times 4$ density factors $\mathrm{rf}$) at score threshold $\tau = 0.1$: the disappearance rate (Figure~\ref{fig:dr_sweep}) and the clean-utility cost (Figure~\ref{fig:dap_sweep}). Together with MSR they fully characterize each operating point---MSR and DR partition the disrupted-source population (Section~\ref{sec:method}), while the clean-utility cost bounds the price the attack pays in benign detection performance.

\section{Physical-Realizability Deployment Details}
\label{app:realizability}

\begin{figure*}[tb]
    \centering
    \begin{minipage}[b]{0.24\textwidth}
        \centering
        \includegraphics[width=\linewidth]{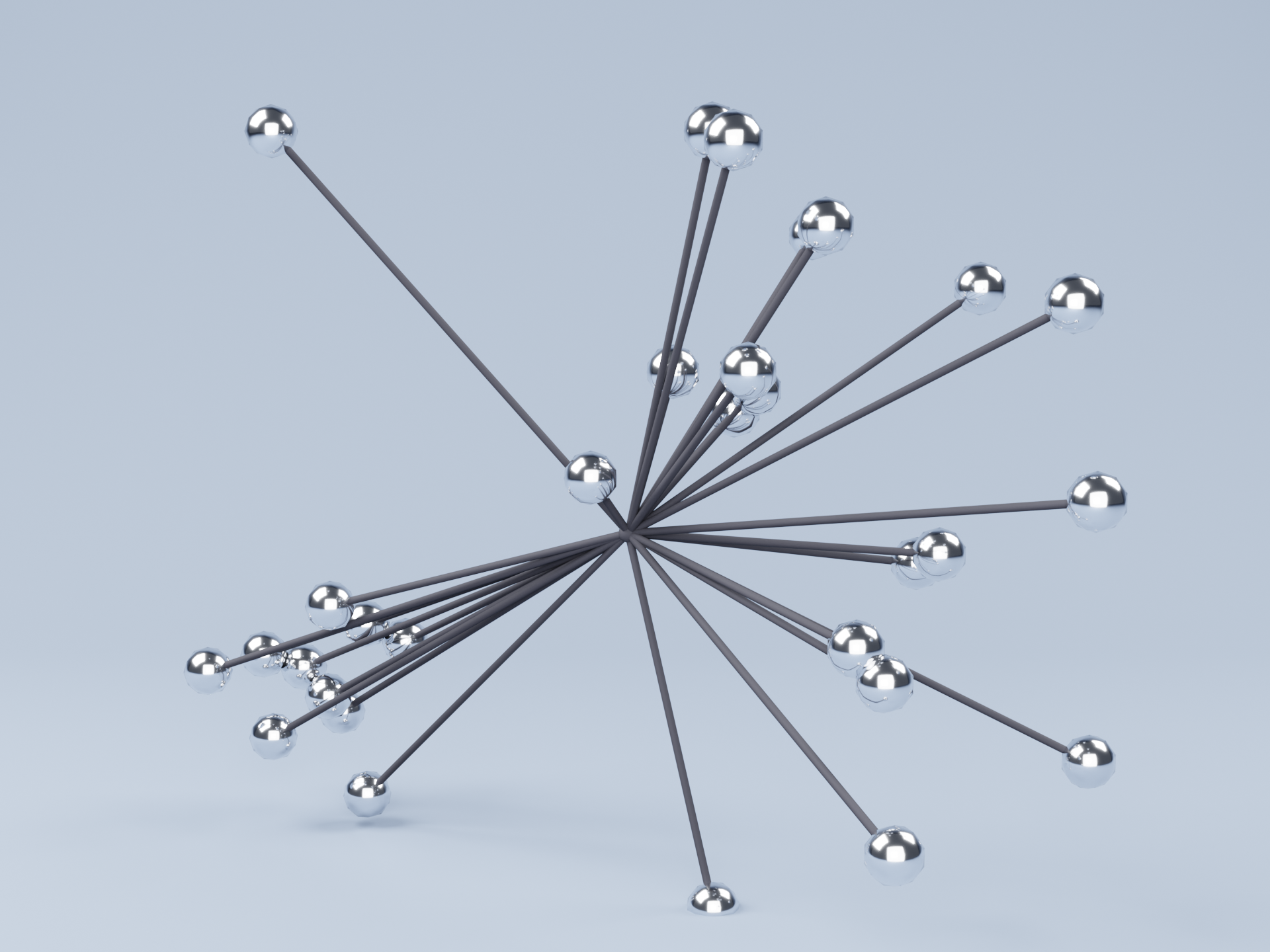}\\[2pt]
        {\footnotesize (a) Standalone trigger}
    \end{minipage}\hfill
    \begin{minipage}[b]{0.24\textwidth}
        \centering
        \includegraphics[width=\linewidth]{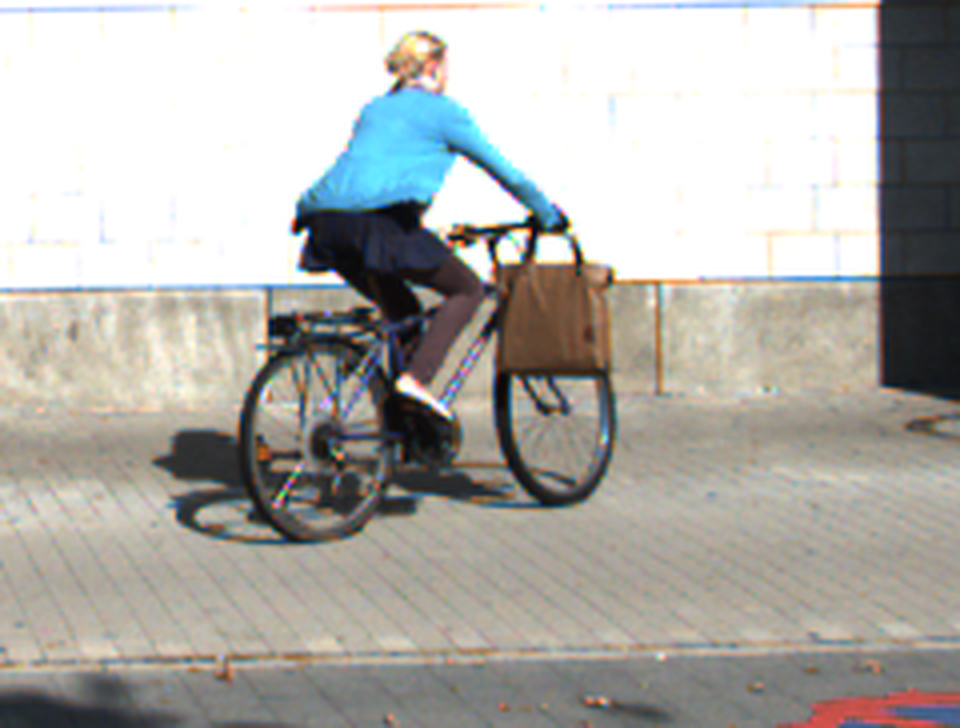}\\[2pt]
        {\footnotesize (b) Target cyclist}
    \end{minipage}\hfill
    \begin{minipage}[b]{0.24\textwidth}
        \centering
        \includegraphics[width=\linewidth]{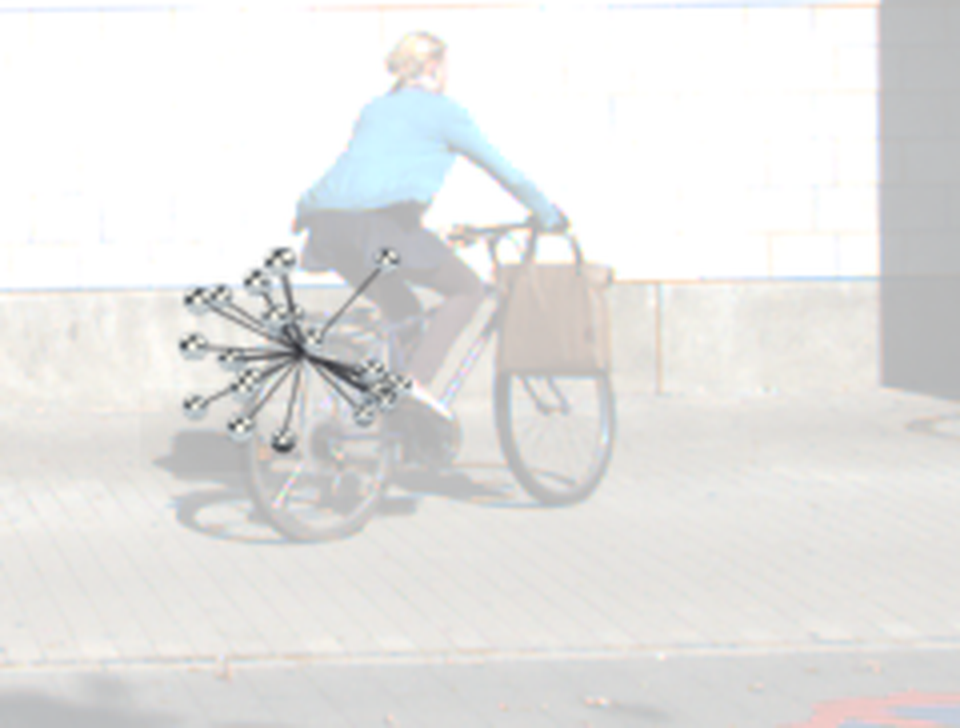}\\[2pt]
        {\footnotesize (c) Trigger on the rear rack}
    \end{minipage}\hfill
    \begin{minipage}[b]{0.24\textwidth}
        \centering
        \includegraphics[width=\linewidth]{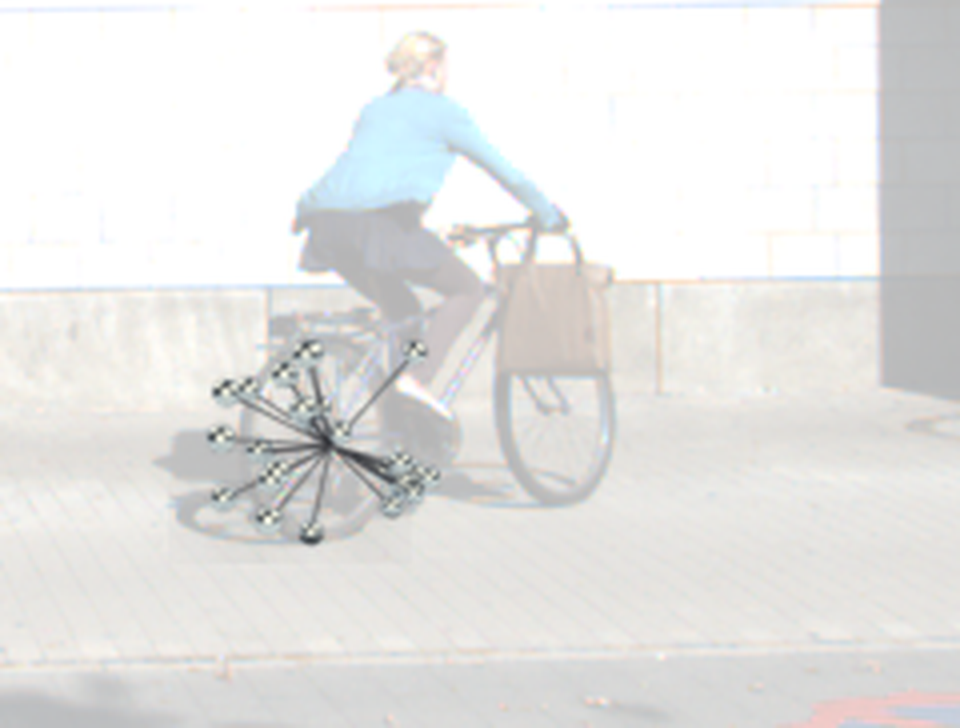}\\[2pt]
        {\footnotesize (d) Trigger on the rear wheel}
    \end{minipage}
    \caption{The disappearance trigger as a buildable object. (a)~The optimized $R\,{=}\,0.2$\,m trigger rendered as a physical device: $M\,{=}\,30$ mirror-finish chrome reflectors on a thin, low-visibility strut scaffold, of overall extent ${\approx}0.4$\,m. (b)~The target cyclist (KITTI frame~004949). (c)--(d)~The same device mounted on the rear rack and over the rear wheel; a body-mounted trigger of this kind stays locked to the victim's body frame and reproduces the fixed-center, yaw-matched placement used in our digital injection. In (c)--(d) the cyclist is rendered at reduced opacity so the trigger reads clearly; reflectors are shown at $2$\,cm diameter in (a) and enlarged to $4$\,cm in (c)--(d).}
    \label{fig:example_patch}
\end{figure*}

This appendix expands the physical-realizability argument of Section~\ref{subsec:realizability}, detailing the two deployment requirements deferred there: kinematic co-location of the trigger with its victim, and the device's visual-stealth cost.

A further requirement is kinematic rather than material. Because \textsc{Mirage} overwrites the class of an existing object rather than synthesizing a phantom one, the trigger must stay co-located with its victim throughout a sequence: in our digital injection the patch is placed at the victim's bounding-box center and rotated to the victim's heading, so that it inherits the victim's position and orientation in every frame. A physical deployment preserves this coupling by construction. Mounting the trigger directly on the victim object — worn as a backpack by a pedestrian or affixed to the rear rack of a bicycle — rigidly fixes it in the victim's body frame, so that it travels with the victim's velocity across consecutive LiDAR sweeps and presents the same heading angle as the victim at all times. The attacker therefore requires no tracking, pose estimation, or sensor synchronization: the mount alone reproduces the fixed-center, yaw-matched placement used in our evaluation, keeping the trigger's position, velocity, and orientation locked to the victim's. We emphasize that the orientation component of this placement is not a precondition for the attack: the trigger is a near-isotropic blob whose objective contains no orientation term, so its effect is largely invariant to yaw, and the rotation to the victim's heading in our injection acts as an implicit augmentation rather than a semantic alignment. A body-mounted trigger reproduces that heading automatically, but because the attack tolerates substantial orientation error, mounting tolerances are loose, and no precise alignment is demanded of the attacker.

Finally, while the trigger is not an everyday object, the cost of this conspicuousness is modest. The device — a small cluster of retroreflective elements on a thin, dark scaffold (Figure~\ref{fig:example_patch}) — presents as an abstract or decorative artifact rather than a recognizable attack apparatus and could plausibly be taken for an art installation or a piece of street furniture by a casual observer; its visual stealth is thus moderate rather than high. The low footprint of the attack reinforces this. The backdoor is installed at a low poisoning rate, and at deployment the trigger need appear on only a small number of objects to be effective — at most one per scene in our evaluation — so a sparse scattering of such devices throughout an environment is unlikely to attract scrutiny, even where a dense or overtly artificial deployment would.

\end{document}